\documentclass{article}
\usepackage{ijcai20}
\usepackage{smile}
\usepackage{times}

\usepackage{soul}
\usepackage{url}
\usepackage[hidelinks]{hyperref}
\usepackage[utf8]{inputenc}
\usepackage[small]{caption}
\usepackage{graphicx}
\usepackage{amsmath}
\usepackage{booktabs}
\def \citep {\cite}
\def \citet {\cite}
\title{Closing the Generalization Gap of Adaptive Gradient Methods in Training \\Deep Neural Networks}

\author{
Jinghui Chen$^1$ \and
Dongruo Zhou$^1$\and
Yiqi Tang$^2$\and
Ziyan Yang$^3$\and
Yuan Cao$^1$\And
Quanquan Gu$^1$\\
\affiliations
$^1$University of California, Los Angeles \\
$^2$Ohio State University\\
$^3$University of Virginia\\
\emails
\{jhchen, drzhou, yuancao, qgu\}@cs.ucla.edu,
tang.1466@osu.edu,
zy3cx@virginia.edu 
}

\begin{document}

\maketitle

\begin{abstract}
Adaptive gradient methods, which adopt historical gradient information to automatically adjust the learning rate, despite the nice property of fast convergence, have been observed to generalize worse than stochastic gradient descent (SGD) with momentum in training deep neural networks. This leaves how to close the generalization gap of adaptive gradient methods an open problem. In this work, we show that adaptive gradient methods such as Adam, Amsgrad, are sometimes ``over adapted''. We design a new algorithm, called \textit{Partially adaptive momentum estimation method}, which unifies the Adam/Amsgrad with SGD by introducing a partial adaptive parameter $p$, to achieve the best from both worlds. 
We also prove the convergence rate of our proposed algorithm to a stationary point in the stochastic nonconvex optimization setting.
Experiments on standard benchmarks show that our proposed algorithm can maintain a fast convergence rate as Adam/Amsgrad while generalizing as well as SGD in training deep neural networks. These results would suggest practitioners 
pick up adaptive gradient methods once again for faster training of deep neural networks.
\end{abstract}

\section{Introduction}\label{sec:intro}
% \vspace{-0.1cm}
Stochastic gradient descent (SGD) is now one of the most dominant approaches for training deep neural networks \citep{goodfellow2016deep}. In each iteration, SGD only performs one parameter update on a mini-batch of training examples. SGD is simple and has been proved to be efficient, especially for tasks on large datasets. 
In recent years, adaptive variants of SGD have emerged and shown their successes for their convenient automatic learning rate adjustment mechanism. Adagrad \citep{duchi2011adaptive} is probably the first along this line of research, and significantly outperforms vanilla SGD in the sparse gradient scenario. Despite the first success, Adagrad was later found to demonstrate degraded performance especially in cases where the loss function is nonconvex or the gradient is dense. Many variants of Adagrad, such as RMSprop \citep{hinton2012neural}, Adam \citep{kingma2014adam}, Adadelta \citep{zeiler2012adadelta}, Nadam \citep{dozat2016incorporating}, were then proposed to address these challenges by adopting exponential moving average rather than the arithmetic average used in Adagrad. This change largely mitigates the rapid decay of learning rate in Adagrad and hence makes this family of algorithms, especially Adam, particularly popular on various tasks. Recently, it has also been observed \citep{reddi2018convergence} that Adam does not converge in some settings where rarely encountered large gradient information quickly dies out due to the ``short momery'' problem of exponential moving average. To address this issue, Amsgrad \citep{reddi2018convergence} has been proposed to keep an extra ``long term memory'' variable to preserve the past gradient information and to correct the potential convergence issue in Adam.
There are also some other variants of adaptive gradient method such as
SC-Adagrad / SC-RMSprop \citep{mukkamala2017variants}, which derives logarithmic regret bounds for strongly convex functions.
%Despite the success of adaptive gradient methods on various optimization tasks, 

On the other hand, people recently found that for largely over-parameterized neural networks, e.g., more complex modern convolutional neural network (CNN) architectures such as VGGNet \citep{he2016deep}, ResNet \citep{he2016deep}, Wide ResNet \citep{zagoruyko2016wide}, DenseNet \citep{huang2017densely}, training with Adam or its variants typically generalizes worse than SGD with momentum, even when the training performance is better. 
In particular, people found that carefully-tuned SGD, combined with proper momentum, weight decay and appropriate learning rate decay schedules, can significantly outperform adaptive gradient algorithms eventually \citep{wilson2017marginal}. As a result, many recent studies train their models with SGD-Momentum \citep{he2016deep,zagoruyko2016wide,huang2017densely,simonyan2014very,ren2015faster,xie2017aggregated,howard2017mobilenets} despite that adaptive gradient algorithms usually converge faster. 
% As a result, even though adaptive gradient methods are relatively easy to tune and converge faster at the early stage, recent advances in designing neural network structures are all reporting their performances by training their models with SGD-momentum \citep{he2016deep,zagoruyko2016wide,huang2017densely,simonyan2014very,ren2015faster,xie2017aggregated,howard2017mobilenets}.
Different from SGD, which adopts a universal learning rate for all coordinates, the effective learning rate of adaptive gradient methods, i.e., the universal base learning rate divided by the second order moment term, is different for different coordinates. Due to the normalization of the second order moment, some coordinates will have very large effective learning rates. To alleviate this problem, one usually chooses a smaller base learning rate for adaptive gradient methods than SGD with momentum.  This makes the learning rate decay schedule less effective when applied to adaptive gradient methods, since a much smaller base learning rate will lead to diminishing effective learning rate for most coordinates after several rounds of decay. We refer to the above phenomenon as the ``small learning rate dilemma'' (see more details in Section \ref{sec:algs}).

With all these observations, a natural question is:

\textit{Can we take the best from both Adam and SGD-Momentum, i.e., design an algorithm that not only enjoys the fast convergence rate as Adam, but also generalizes as well as SGD-Momentum? }

In this paper, we answer this question affirmatively. We close the generalization gap of adaptive gradient methods by proposing a new algorithm, called \textbf{p}artially \textbf{ada}ptive \textbf{m}omentum estimation (Padam) method, which unifies Adam/Amsgrad with SGD-Momentum to achieve the best of both worlds, by a partially adaptive parameter. The intuition behind our algorithm is: by controlling the degree of adaptiveness, the base learning rate in Padam does not need to be as small as other adaptive gradient methods. Therefore, it can maintain a larger learning rate while preventing the gradient explosion. 
We note that there exist several studies \citep{manzil2018adaptive,loshchilov2017fixing,Luo2019AdaBound} that also attempted to address the same research question. In detail, 
Yogi \citep{manzil2018adaptive} studied the effect of adaptive denominator constant $\epsilon$ and minibatch size in the convergence of adaptive gradient methods. 
AdamW \citep{loshchilov2017fixing} proposed to fix the weight decay regularization in Adam by decoupling the weight decay from the gradient update and this improves the generalization performance of Adam. AdaBound \citep{Luo2019AdaBound} applies dynamic bound of learning rate on Adam and make them smoothly converge to a constant final step size as in SGD. Our algorithm is very different from Yogi, AdamW and AdaBound. Padam is built upon a simple modification of Adam without extra complicated algorithmic design and it comes with a rigorous convergence guarantee in the nonconvex stochastic optimization setting.

We highlight the main contributions of our work as follows:
% \vspace{-0.2cm}
% \begin{itemize}
% \setlength{\itemindent}{0em}
% \begin{itemize}

% \item 
$\bullet$ We propose a novel and simple algorithm Padam with a partially adaptive parameter, which resolves the ``small learning rate dilemma'' for adaptive gradient methods and allows for faster convergence, hence closing the gap of generalization.

$\bullet$ We provide a convergence guarantee for Padam in nonconvex optimization.
% We provide a convergence analysis of Padam in the convex setting, based on the analysis of \citet{kingma2014adam, reddi2018convergence}, and prove a data-dependent regret bound. 
Specifically, we prove that the convergence rate of Padam to a stationary point for stochastic nonconvex optimization is  
% \vspace{-0.05cm}
\begin{align}\label{eq:Gu9999}
    O\bigg(\frac{ d^{1/2} }{T^{3/4 - s/2}} + \frac{d}{T}\bigg),
\end{align}
where $s$ characterizes the
growth rate of the cumulative stochastic gradient $\gb_{1:T,i} = [g_{1,i},g_{2,i},\ldots,g_{T,i}]^\top$ ($\gb_1,\ldots,\gb_T$ are the stochastic gradients)
and $0 \leq s \leq 1/2$. When the stochastic gradients are sparse, i.e., $s < 1/2$, \eqref{eq:Gu9999} is strictly better than the convergence rate of SGD in terms of the rate of $T$.

% where $\gb_1,\ldots,\gb_T$ are the stochastic gradients and $\gb_{1:T,i} = [g_{1,i},g_{2,i},\ldots,g_{T,i}]^\top$. When the stochastic gradients are $\ell_\infty$-bounded, \eqref{eq:Gu9999} matches the convergence rate of SGD in terms of the rate of $T$.

% \item
$\bullet$ We also provide thorough experiments about our proposed Padam method on training modern deep neural architectures. We empirically show that Padam achieves the fastest convergence speed while generalizing as well as SGD with momentum. These results suggest that practitioners should pick up adaptive gradient methods once again for faster training of deep neural networks.

$\bullet$ Last but not least, compared with the recent work on adaptive gradient methods, such as Yogi \citep{manzil2018adaptive}, AdamW \citep{loshchilov2017fixing}, AdaBound \citep{Luo2019AdaBound}, our proposed Padam achieves better generalization performance than these methods in our experiments.

    % \item Last but not least, our partially adaptive solution is not limited to Adam/Amsgrad. We believe that it can also be applied to other adaptive gradient methods and help close their generalization gap. \CC{This point is very vague}
    
% \end{itemize}
% \vspace{-0.4cm}

% \vspace{-0.1cm}
\subsection{Additional Related Work}\label{sec:re}
% \vspace{-0.1cm}
Here we review additional related work that is not covered before.
% There are only several studies closely related to improving the generalization performance of Adam.
\citet{zhang2017normalized} proposed a normalized direction-preserving Adam (ND-Adam), which changes the adaptive terms from individual dimensions to the whole gradient vector. 
% This makes it more like a SGD-momentum with learning rate adjusted in each iteration, rather than an adaptive gradient algorithm. The empirical result also shows that its performance resembles SGD with momentum.
\citet{keskar2017improving} proposed to improve the generalization performance by switching from Adam to SGD. 
% Yet the empirical result shows that it actually sacrifices some of the convergence speed for better generalization rather than achieving the best from both worlds.
% Also deciding the switching learning rate and the best switching point requires extra efforts on parameter tuning since they can be drastically different for different tasks according to the paper. 
On the other hand, despite the great successes of adaptive gradient methods for training deep neural networks, the convergence guarantees for these algorithms are still understudied. Most convergence analyses of adaptive gradient methods are restricted to online convex optimization \citep{duchi2011adaptive,kingma2014adam,mukkamala2017variants,reddi2018convergence}. 
A few recent attempts have been made to analyze adaptive gradient methods for stochastic nonconvex optimization. 
More specifically,
\citet{basu2018convergence} proved the convergence rate of RMSProp and Adam when using deterministic gradient rather than stochastic gradient. \citet{li2018convergence} analyzed convergence rate of AdaGrad under both convex and nonconvex settings but did not consider more complicated Adam-type algorithms. 
\citet{ward2018adagrad} also proved the convergence rate of AdaGrad under both convex and nonconvex settings without considering the effect of stochastic momentum. \citet{chen2018on} provided a convergence analysis for a class of Adam-type algorithms for nonconvex optimization. \citet{zou2018convergence} analyzed the convergence rate of AdaHB and AdaNAG, two modified version of AdaGrad with the use of momentum. %These two algorithms are different from Padam and AMSGrad, and their results not directly comparable to ours. 
\citet{liu2019towards} proposed Optimistic Adagrad and showed its convergence in  non-convex non-concave min-max optimization. 
However, none of these results are directly applicable to Padam. Our convergence analysis in Section \ref{sec:theory} is quite general and implies the convergence rate of AMSGrad for nonconvex optimization. 
In terms of learning rate decay schedule, \cite{wu2018understanding} studied the learning rate schedule via short-horizon bias. \cite{xu2016accelerate,davis2019stochastic} analyzed the convergence of stochastic algorithms with geometric learning rate decay. \cite{ge2019step} studied the learning rate schedule for quadratic functions.

The remainder of this paper is organized as follows: in Section \ref{sec:related}, we briefly review existing adaptive gradient methods. We present our proposed algorithm in Section \ref{sec:algs}, and the main theory in Section \ref{sec:theory}.  In Section \ref{sec:exp}, we compare the proposed algorithm with existing algorithms on modern neural network architectures on  benchmark datasets. %We also briefly review the related work in Section \ref{sec:re}. 
Finally, we conclude this paper and point out the future work in Section \ref{sec:conclusion}. 
 
% \vspace{-0.3cm}
\paragraph{Notation:} Scalars are denoted by lower case letters, vectors by lower case bold face letters, and matrices by upper case bold face letters. For a vector $\xb \in \RR^d$, we denote the $\ell_2$ norm of $\xb$ by $\| \xb \|_2 = \sqrt{\sum_{i=1}^d x_i^2}$, the $\ell_\infty$ norm of $\xb$ by $ \|\xb\|_\infty = \max_{i=1}^d |x_i|$. For a sequence of vectors $\{\xb_j\}_{j=1}^t$, we denote by $x_{j,i}$ the $i$-th element in $\xb_j$. We also denote $\xb_{1:t,i} = [x_{1,i},\ldots, x_{t,i}]^\top$. With slight abuse of notation, for two vectors $\ab$ and $\bbb$, we denote $\ab^2$ as the element-wise square, ${\ab}^p$ as the element-wise power operation, $\ab / \bbb$ as the element-wise division and $\max(\ab, \bbb)$ as the element-wise maximum. 
 %\CC{positive semidefinite, or positive definite, if positive definite, you need to use $\cS^d_{++}$} 
We denote by $\diag(\ab)$ a diagonal matrix with diagonal entries $a_1,\ldots,a_d$.
% Let $\cS^d_{++}$ be the set of all positive definite $d \times d$ matrices. 
% We denote by $\Pi_{\cX, \Ab}(\bbb)$ the projection of $\bbb$ onto a convex set $\cX$, i.e., $\argmin_{\ab \in \cX} \|\Ab^{1/2}(\ab - \bbb)\|_2$ for $\bbb \in \RR^d, \Ab \in \cS^d_{++}$.
Given two sequences $\{a_n\}$ and $\{b_n\}$, we write $a_n=O(b_n)$ if there exists a positive constant $C$ such that $a_n \leq C b_n$ and $a_n = o(b_n)$ if $a_n/b_n\rightarrow 0$ as $n\rightarrow \infty$. Notation $\tilde{O}(\cdot)$ hides logarithmic factors.

% \vspace{-0.1cm}
\section{Review of Adaptive Gradient Methods}\label{sec:related}
% \vspace{-0.1cm}
Various adaptive gradient methods have been proposed in order to achieve better performance on various stochastic optimization tasks. Adagrad \citep{duchi2011adaptive} is among the first methods with adaptive learning rate for each individual dimension, which motivates the research on adaptive gradient methods in the machine learning community. In detail, Adagrad\footnote{The formula is equivalent to the one from the original paper \citep{duchi2011adaptive} after simple manipulations.} adopts the following update form:
% \vspace{-0.2cm}
\begin{align*} 
    \xb_{t+1} = \xb_{t} - \alpha_t \frac{\gb_t}{\sqrt{\vb_t}}, \mbox{ where } \vb_t = \frac{1}{t} \sum_{j=1}^t \gb_j^2,
\end{align*}
% \vspace{-0.1cm}
where $\gb_t$ stands for the stochastic gradient $\nabla f_t(\xb_t)$, and $\alpha_t = \alpha / \sqrt{t}$ is the step size. In this paper, we call $\alpha_t$ \textit{base learning rate}, which is the same for all coordinates of $\xb_t$, and we call $\alpha_t/\sqrt{v_{t,i}}$ \textit{effective learning rate} for the $i$-th coordinate of $\xb_t$, which varies across the coordinates. Adagrad is proved to enjoy a huge gain in terms of convergence especially in sparse gradient situations. Empirical studies also show a performance gain even for non-sparse gradient settings. RMSprop \citep{hinton2012neural} follows the idea of adaptive learning rate and it changes the arithmetic averages used for $\vb_t$ in Adagrad to exponential moving averages. Even though RMSprop is an empirical method with no theoretical guarantee, the outstanding empirical performance of RMSprop raised people's interests in exponential moving average variants of Adagrad. Adam \citep{kingma2014adam}\footnote{Here for simplicity and consistency, we ignore the bias correction step in the original paper of Adam. Yet adding the bias correction step will not affect the argument in the paper.} is the most popular exponential moving average variant of Adagrad. It combines the idea of RMSprop and momentum acceleration, and takes the following update form:
% \vspace{-0.1cm}
\begin{align*} 
    &\xb_{t+1} = \xb_{t} - \alpha_t \frac{\mb_t}{\sqrt{\vb_t}} \mbox{ where } \\ &\mb_t = \beta_1 \mb_{t-1} + (1- \beta_1) \gb_t, \vb_t = \beta_2 \vb_{t-1} + (1- \beta_2) \gb_t^2.
\end{align*}
Adam also requires $\alpha_t = \alpha/\sqrt{t}$ for the sake of convergence analysis. In practice, any decaying step size or even constant step size works well for Adam. Note that if we choose $\beta_1 = 0$, Adam basically reduces to RMSprop.  \citet{reddi2018convergence} identified a non-convergence issue in Adam. Specifically, Adam does not collect long-term memory of past gradients and therefore the effective learning rate could be increasing in some cases. They proposed a modified algorithm namely Amsgrad. More specifically, Amsgrad adopts an additional step to ensure the decay of the effective learning rate $\alpha_t/\sqrt{\hat \vb_t}$, and its key update formula is as follows: %\footnote{The true learning rate here refers to the learning rate after adaptive adjustment.}:
% \vspace{-0.1cm}
\begin{align*} 
    \xb_{t+1} = \xb_{t} - \alpha_t \frac{\mb_t}{\sqrt{\hat\vb_t}}, \mbox{ where }  \hat\vb_t = \max(\hat \vb_{t-1}, \vb_t),
\end{align*}
$\mb_t$ and $\vb_t$ are the same as Adam. By introducing the $\hat\vb_t$ term, \citet{reddi2018convergence} corrected some mistakes in the original proof of Adam and proved an $O(1/\sqrt{T})$ convergence rate of Amsgrad for convex optimization. Note that all the theoretical guarantees on adaptive gradient methods (Adagrad, Adam, Amsgrad) are only proved for convex functions.

% \vspace{-0.2cm}
\section{The Proposed Algorithm}\label{sec:algs}
% \vspace{-0.2cm}
In this section, we propose a new algorithm for bridging the generalization gap for adaptive gradient methods.
Specifically, we introduce a partial adaptive parameter $p$ to control the level of adaptiveness of the optimization procedure.
The proposed algorithm is displayed in Algorithm \ref{alg:Padam}. 

% \vspace{-0.2cm}
\begin{algorithm}[t]
	\caption{Partially adaptive momentum estimation method (Padam)}
	\label{alg:Padam}
	\begin{algorithmic}
		\STATE \textbf{input:} initial point $\xb_1 \in \cX$; step sizes $\{\alpha_t\}$; adaptive parameters $\beta_1, \beta_2$, $p \in (0, 1/2]$ 
		\STATE set $\mb_0 = \zero$, $\vb_0 = \zero$, $\hat\vb_0 = \zero$
 		\FOR {$t = 1,\ldots, T$}
		     \STATE $\gb_t=\nabla f(\xb_t, \xi_t)\;$
		     \STATE $\mb_t = \beta_1\mb_{t-1} + (1 - \beta_1)\gb_t$
		     \STATE $\vb_t = \beta_2\vb_{t-1} + (1 - \beta_2)\gb_t^2$
		     \STATE $\hat\vb_t = \max(\hat\vb_{t-1}, \vb_t)$
		     \STATE $\xb_{t+1} =   \xb_{t} - \alpha_t \cdot   \mb_t / {\hat \vb_t}^{p}   $
		\ENDFOR     
	    \STATE \textbf{Output:} Choose $\xb_{\text{out}}$ from $\{\xb_t\}, 2\leq t \leq T$ with probability $\alpha_{t-1}/\big(\sum_{i=1}^{T-1}\alpha_i)$
	\end{algorithmic}
\end{algorithm}
% \vspace{-0.2cm}
 
In Algorithm \ref{alg:Padam}, $\gb_t$ denotes the stochastic gradient and $\hat\vb_t$ can be seen as a moving average over the second order moment of the stochastic gradients. 
As we can see from Algorithm \ref{alg:Padam}, the key difference between Padam and Amsgrad \citep{reddi2018convergence} is that: while $\mb_t$ is still the momentum as in Adam/Amsgrad, it is now ``partially adapted'' by the second order moment.
% , i.e.,
% \begin{align*}\tag{Padam}
%     \xb_{t+1} = \xb_{t} - \alpha_t \frac{\mb_t}{\hat \vb_t^p}, \mbox{ where }  \hat\vb_t = \max(\hat \vb_{t-1}, \vb_t).
% \end{align*}
We call $p \in [0,1/2]$ the partially adaptive parameter. Note that $1/2$ is the largest possible value for $p$ and a larger $p$ will result in non-convergence in the proof (see the proof details in the supplementary materials). 
When $p \to 0$, Algorithm \ref{alg:Padam} reduces to SGD with momentum\footnote{The only difference between Padam with $p = 0$ and SGD-Momentum is an extra constant factor $(1 -\beta_1)$, which can be moved into the learning rate such that the update rules for these two algorithms are identical.} and when $p  = 1/2$, Algorithm \ref{alg:Padam} is exactly Amsgrad. Therefore, Padam indeed unifies Amsgrad and SGD with momentum.

% \CC{Note that even though in Algorithm \ref{alg:Padam}, the choice of $\alpha_t$ covers different choices of learning rate decay schedule, the main focus of this paper is not about finding the best learning rate decay schedule\footnote{\CC{As a result, we simply fix the learning rate decay schedule for all method in the experiments.}}, but designing a new algorithm to control the adaptiveness for better empirical generalization result.} 

With the notations defined above, we are able to formally explain the ``small learning rate dilemma''. In order to make things clear, we first emphasize the relationship between adaptiveness and learning rate decay. We refer the actual learning rate applied to $\mb_t$ as the effective learning rate, i.e., $\alpha_t/\hat \vb_t^p$. % where $h(\cdot)$ is a function of $\hat \vb_t$. Clearly, $h(\hat \vb_t) = \sqrt{\hat \vb_t}$ in Amsgrad and $h(\hat \vb_t) = {\hat \vb_t}^p$ in Padam.
% In Amsgrad, we have $h(\hat \vb_t) = \sqrt{\hat \vb_t}$ while in Padam, we have $h(\hat \vb_t) = {\hat \vb_t}^p$. 
Now suppose that a learning rate decay schedule is applied to $\alpha_t$. If $p$ is large, then at early stages, the effective learning rate $\alpha_t/{\hat v_{t,i}}^p$ could be fairly large for certain coordinates with small $\hat v_{t,i}$ value\footnote{The coordinate $\hat v_{t,i}$'s are much less than $1$ for most commonly used network architectures. 
% See Figure \ref{fig:max_min_v} in the supplementary materials.
}. To prevent those coordinates from overshooting we need to enforce a smaller $\alpha_t$, and therefore the base learning rate must be set small \citep{keskar2017improving,wilson2017marginal}. %(This is the reason why the base learning rate of adaptive gradient methods needs to be much smaller than that of SGD-momentum \citep{keskar2017improving, wilson2017marginal} when training deep neural networks.)
As a result, after several rounds of decaying, the learning rates of the adaptive gradient methods are too small to make any significant progress in the training process\footnote{This does not mean the learning rate decay schedule weakens adaptive gradient method. On the contrary, applying the learning rate decay schedule still gives performance boost to the adaptive gradient methods in general but this performance boost is not as significant as SGD + momentum.}. We call this phenomenon ``small learning rate dilemma''. It is also easy to see that the larger $p$ is, the more severe ``small learning rate dilemma'' is.
This suggests that intuitively, we should consider using Padam with a proper adaptive parameter $p$, and choosing $p<1/2$ can potentially make Padam suffer less from the ``small learning rate dilemma'' than Amsgrad, which justifies the range of $p$ in Algorithm \ref{alg:Padam}. %To back up this intuition, in Figure \ref{fig:max_min_v}, we present the performance comparison of Padam with different choices of $p$. 
% which will enable us to adopt a larger base learning rate $\alpha_t$ to avoid the ``small learning rate dilemma''. Since the coordinate $\hat v_{t,i}$'s are much less than $1$ for most commonly used network architectures (See Figure \ref{fig:max_min_v} in the supplementary materials), in order to decrease the effective learning rate of Padam (i.e., $\alpha_t/({v_{t,i}}^p) < \alpha_t/\sqrt{v_{t,i}}$), we should choose $p$ to be less than $1/2$. This again justifies the range of $p$ in Algorithm \ref{alg:Padam} from the empirical perspective.
We will show in our experiments (Section \ref{sec:exp}) that Padam with $p<1/2$ can adopt an equally large base learning rate as SGD with momentum.

Note that even though in Algorithm \ref{alg:Padam}, the choice of $\alpha_t$ covers different choices of learning rate decay schedule, the main focus of this paper is not about finding the best learning rate decay schedule, but designing a new algorithm to control the adaptiveness for better empirical generalization result. In other words, our focus is not on $\alpha_t$, but on $\hat \vb_t$. For this reason, we simply fix the learning rate decay schedule for all methods in the experiments to provide a fair comparison for different methods.

\begin{figure*}[h]
  \centering
  \subfigure[CIFAR-10]{\includegraphics[width=0.35\textwidth]{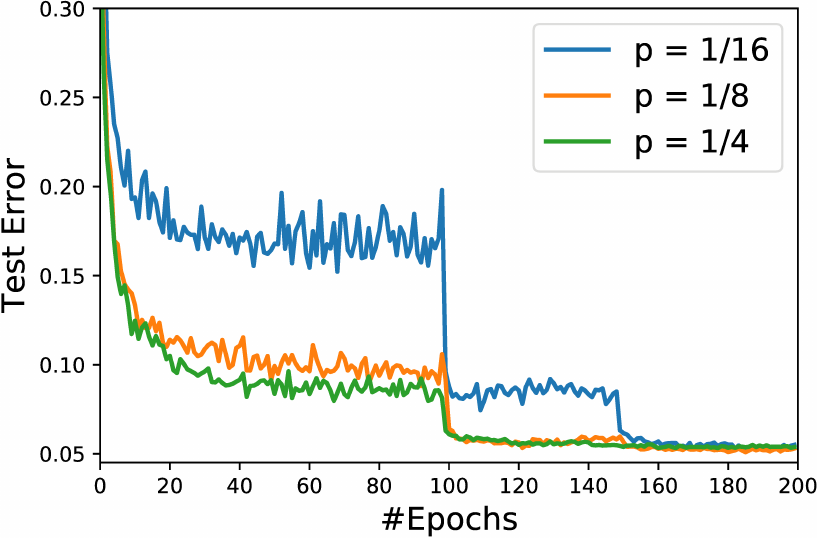}}
  \subfigure[CIFAR-100]{\includegraphics[width=0.35\textwidth]{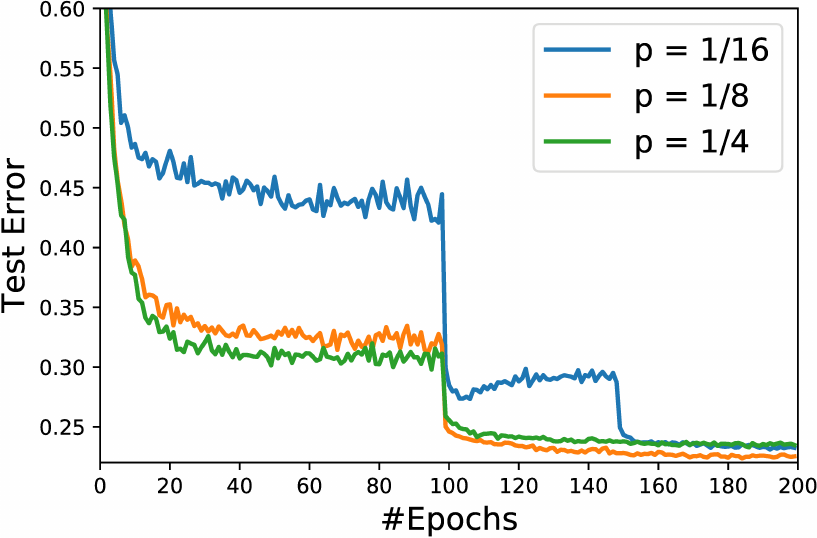}}
  \caption{Performance comparison of Padam with different choices of $p$ for training ResNet on (a) CIFAR-10  and (b) CIFAR-100 datasets.}
  \label{fig:exp_p}
\end{figure*} 
 
Figure \ref{fig:exp_p} shows the comparison of test error performances under the different partial adaptive parameter $p$ for ResNet on both CIFAR-10 and CIFAR-100 datasets. 
We can observe that a larger $p$ will lead to fast convergence at early stages and worse generalization performance later, while a smaller $p$ behaves more like SGD with momentum: slow in early stages but finally catch up. With a proper choice of $p$ (e.g., $1/8$ in this case), Padam can obtain the best of both worlds.  
Note that besides Algorithm \ref{alg:Padam}, our partially adaptive idea can also be applied to other adaptive gradient methods such as Adagrad, Adadelta, RMSprop, AdaMax \citep{kingma2014adam}. For the sake of conciseness, we do not list the partially adaptive versions for other adaptive gradient methods here. We also would like to comment that Padam is totally different from the $p$-norm generalized version of Adam in \citet{kingma2014adam}, which induces AdaMax method when $p \to \infty$. In their case, $p$-norm is used to generalize $2$-norm of their current and past gradients while keeping the scale of adaptation unchanged. In sharp contrast, we intentionally change (reduce) the scale of the adaptive term in Padam to get better generalization performance.
% Finally, note that in Algorithm \ref{alg:Padam} we remove the bias correction step used in the original Adam paper following \citet{reddi2018convergence}. Nevertheless, our arguments and theory are applicable to the bias correction version as well.
 
% \vspace{-0.1cm}
\section{Convergence Analysis of the Proposed Algorithm}\label{sec:theory}
% \vspace{-0.1cm}

In this section, we establish the convergence theory of Algorithm \ref{alg:Padam} in the stochastic nonconvex optimization setting, i.e., we aim at solving the following stochastic nonconvex optimization problem
\begin{align}%\label{eq:intro_1}
    \min_{\xb \in \RR^d} f(\xb):=\EE_\xi \big[f(\xb;\xi)\big],\notag
\end{align}
where $\xi$ is a random variable satisfying certain distribution, $f(\xb; \xi):\RR^d \rightarrow \RR$ is a $L$-smooth nonconvex function. In the stochastic setting, one cannot directly access the full gradient of $f(\xb)$. Instead, one can only get unbiased estimators of the gradient of $f(\xb)$, which is $\nabla f(\xb;\xi)$. This setting has been studied in \citet{ghadimi2013stochastic,ghadimi2016accelerated}. 
We first introduce the following assumptions. 

\begin{assumption}[Bounded Gradient]\label{as:1}
$f(\xb) = \EE_\xi f(\xb ; \xi)$ has $G_\infty$-bounded stochastic gradient. That is, for any $\xi$, we assume that $\|\nabla f(\xb ; \xi)\|_{\infty} \leq G_{\infty}$.
% \begin{align*}
%   \|\nabla f(\xb ; \xi)\|_{\infty} \leq G_{\infty}.
%  \end{align*}
\end{assumption}

It is worth mentioning that Assumption~\ref{as:1} is slightly weaker than the $\ell_2$-boundedness assumption $\|\nabla f(\xb ; \xi)\|_{2} \leq G_{2}$ used in \cite{reddi2016stochastic,chen2018on}. Since $\|\nabla f(\xb ; \xi)\|_{\infty} \leq \|\nabla f(\xb ; \xi)\|_{2} \leq \sqrt{d}\|\nabla f(\xb ; \xi)\|_{\infty}$, the $\ell_2$-boundedness assumption implies Assumption~\ref{as:1} with $G_\infty = G_2$. Meanwhile, $G_\infty$ will be tighter than $G_2$ by a factor of $\sqrt{d}$ when each coordinate of $\nabla f(\xb ; \xi)$ almost equals to each other.

\begin{assumption}[$L$-smooth]\label{as:2}
$f(\xb) = \EE_\xi f(\xb ; \xi)$ is $L$-smooth: for any $\xb, \yb \in \RR^d$, it satisfied that
% \vspace{-0.1cm}
% \begin{align*}
   $\big|f(\xb) - f(\yb)- \langle \nabla f(\yb), \xb-\yb \rangle\big| \leq \frac{ L}{2}\|\xb-\yb\|_2^2.$
% \end{align*}
\end{assumption}
% \vspace{-0.2cm}
Assumption~\ref{as:2} is frequently used in analysis of gradient-based algorithms. It is equivalent to the $L$-gradient Lipschitz condition, which is often written as $\|\nabla f(\xb)-\nabla f(\yb)\|_2\le L\|\xb-\yb\|_2$.
Next we provide the main convergence rate result for our proposed algorithm.
The detailed proof can be found in the longer version of this paper.

\begin{theorem}\label{eq:Gu0000}
In Algorithm \ref{alg:Padam}, suppose that $p\in[0,1/2]$, $\beta_1 < \beta_2^{2p}$, $\alpha_t = \alpha$ and $\|\gb_{1:T,i}\|_2 \leq  G_{\infty}T^s$ for $t=1,\ldots,T$, $0 \leq s \leq 1/2$, under Assumptions~\ref{as:1} and \ref{as:2}, let  $\Delta f=f(\xb_1) - \inf_{\xb} f(\xb)$, for any $q\in[\max\{0,4p-1\},1]$, the output $\xb_{\text{out}}$ of Algorithm \ref{alg:Padam} satisfies that
% \vspace{-0.2cm}
{\small
\begin{align}
    % \EE\Big[\big\|\nabla f(\xb_{\text{out}})\big\|_2^2\Big] &\leq \frac{M_1}{T\alpha} + \frac{M_2 d}{T}+ \frac{M_3\alpha d^q}{T^{(1-q)/2}}\EE \bigg(\sum_{i=1}^d  \|\gb_{1:T,i}\|_2\bigg)^{1-q},
    \EE\Big[\big\|\nabla f(\xb_{\text{out}})\big\|_2^2\Big]  
    &\leq \frac{M_1}{T\alpha} + \frac{M_2 d}{T}+ \frac{M_3\alpha d G_{\infty}^{1-q}}{T^{(1-q)(1/2 - s)}}, \label{mainbound_0} 
\end{align}}
where %$\{M_i\}_{i=1}^3$ are defined as follows
% \vspace{-0.2cm}
{\small
\begin{align*}
    M_1 &= 2G_\infty^{2p}\Delta f, \ M_2 = \frac{4G_\infty^{2+2p}\EE\big\| \hat{\vb}_{1}^{-p}\big\|_1}{d(1-\beta_1)} + 4G_\infty^2, \\
    M_3 &= \frac{4LG_\infty^{1+q-2p}}{(1-\beta_2)^{2p}} + \frac{8LG_\infty^{1+q-2p}(1-\beta_1)}{(1-\beta_2)^{2p}(1-\beta_1/\beta_2^{2p})}\bigg(\frac{\beta_1}{1 - \beta_1}\bigg)^2.
\end{align*}}
\end{theorem}
\begin{remark}
From Theorem \ref{eq:Gu0000}, we can see that $M_1$ and $M_3$ are independent of the number of iterations $T$ and dimension $d$. In addition, if $\| \hat{\vb}_{1}^{-1} \|_{\infty} =O(1)$, it is easy to see that $M_2$ also has an upper bound that is independent of $T$ and $d$. 
$s$ characterizes the
growth rate condition \citep{liu2019towards} of the cumulative stochastic gradient $\gb_{1:T,i}$. In the worse case, $s = 1/2$, while in practice when the stochastic gradients are sparse, $s < 1/2$.
\end{remark}
The following corollary 
% is a special case of Theorem~\ref{eq:Gu0000} when $p\in [0,1/4]$ and $q=0$.
simplifies the result of Theorem \ref{eq:Gu0000} by choosing $q=0$ under the condition $p\in [0,1/4]$.

%In the following corollary, we show that our bound implies that Padam achieves faster convergence when $p$ is located in $[0,1/4]$.
\begin{corollary}\label{cl:1}
Under the same conditions in Theorem \ref{eq:Gu0000}, if $p \in [0,1/4]$, Padam's output satisfies
{\small\begin{align}
    \EE\Big[\big\|\nabla f(\xb_{\text{out}})\big\|_2^2\Big]  \leq \frac{M_1}{T\alpha} + \frac{M_2 d}{T}+ \frac{M_3'\alpha d G_{\infty}}{T^{1/2 - s}}  ,\label{mainbound_1}
\end{align}}
where $M_1$ and $M_2$ and $\Delta f$ are the same as in Theorem \ref{eq:Gu0000}, and $M_3'$ is defined as follows:
{\small\begin{align}
    M_3'  = \frac{4LG_\infty^{1-2p}}{(1-\beta_2)^{2p}} + \frac{8LG_\infty^{1-2p}(1-\beta_1)}{(1-\beta_2)^{2p}(1-\beta_1/\beta_2^{2p})}\bigg(\frac{\beta_1}{1 - \beta_1}\bigg)^2.\notag
\end{align}}
%\CC{do we need the following second part?}
\iffalse
For $p \in (1/4,1/2]$, we set $q$ in Theorem \ref{eq:Gu0000} to $4p-1$, then with the same assumption in Theorem \ref{eq:Gu0000}, we have
\begin{align}
    \EE\Big[\big\|\nabla f(\xb_{\text{out}})\big\|_2^2\Big] \leq \frac{M_1}{\sqrt{T}} + \frac{M_2 d}{T}+ \frac{d^{4p-1}M_3''}{T^{3/2-2p}}\EE \bigg(\sum_{i=1}^d  \|\gb_{1:T,i}\|_2\bigg)^{2-4p}, \notag
\end{align}
where $M_1$ and $M_2$ are variables defined by \eqref{defm1} and \eqref{defm2} in Theorem \eqref{eq:Gu0000}, 
\begin{align}
    M_3'  = \frac{4LG_\infty^{2p}}{(1-\beta_2)^{2p}} + \frac{8LG_\infty^{2p}(1-\beta_1)}{(1-\beta_2)^{2p}(1-\beta_1/\beta_2^{2p})}\bigg(\frac{\beta_1}{1 - \beta_1}\bigg)^2.\notag 
\end{align}
\fi
\end{corollary}

% \begin{remark}\label{remark:sumgnorm}
% Corollary~\ref{cl:1} simplifies the result of Theorem \ref{eq:Gu0000} by choosing $q=0$ under the condition $p\in [0,1/4]$. 
% We remark that this choice of $q$ is optimal in an important special case studied in \citet{duchi2011adaptive,reddi2018convergence}: when the gradient vectors are sparse, we assume that $\sum_{i=1}^d  \|\gb_{1:T,i}\|_2\ll \sqrt{dT}$. Then for $q>0$, it follows that
% {\small\begin{align}
%     \frac{\sum_{i=1}^d  \|\gb_{1:T,i}\|_2}{T} \ll \frac{d^q \big(\sum_{i=1}^d  \|\gb_{1:T,i}\|_2\big)^{1-q}}{T^{1-q/2}}. \label{strict}
% \end{align}}
% \eqref{strict} implies that the upper bound provided by \eqref{mainbound_1} is strictly better than \eqref{mainbound_0} with $q>0$. Therefore when the gradient vectors are sparse, Padam achieves faster convergence when $p \in [0,1/4]$.
% \end{remark}
\begin{remark}\label{rr1}
We show the convergence rate under optimal choice of step size $\alpha$. If  
{\small\begin{align}
    \alpha = \Theta\big(d^{1/2}T^{1/4 + s/2} \big)^{-1},\notag
\end{align}}
then by \eqref{mainbound_1}, we have
{\small\begin{align}
    \EE\Big[\big\|\nabla f(\xb_{\text{out}})\big\|_2^2\Big] = O\bigg(\frac{d^{1/2} }{T^{3/4 - s/2}} + \frac{d}{T}\bigg). \label{step1}
\end{align}}
Note that the convergence rate given by \eqref{step1} is related to $s$.  
In the worst case when $s = 1/2$, we have
$\EE[\|\nabla f(\xb_{\text{out}})\|_2^2] = O\big(\sqrt{d/T}+d/T\big),$
which matches the rate $O(1/\sqrt{T})$ achieved by nonconvex SGD \citep{ghadimi2016accelerated}, considering the dependence of $T$. 
When the stochastic gradients $\gb_{1:T,i}$, $i=1,\ldots,d$ are sparse, i.e., $s < 1/2$, the convergence rate in \eqref{step1} is strictly better than the convergence rate of nonconvex SGD \citep{ghadimi2016accelerated}.
% When the stochastic gradients $\gb_{1:T,i}$, $i=1,\ldots,d$ are sparse, we have $\sum_{i=1}^d  \|\gb_{1:T,i}\|_2\ll \sqrt{dT}$ (\citet{duchi2011adaptive}). 
\end{remark}

% \begin{remark}\label{rr2}
% If we set $\alpha = 1/\sqrt{T}$ which is not related to $\sum_{i=1}^d  \|\gb_{1:T,i}\|_2$, then \eqref{mainbound_1} suggests that 
% {\small\begin{align*}
%     \EE\Big[\big\|\nabla f(\xb_{\text{out}})\big\|_2^2\Big] = O\bigg(\frac{1}{\sqrt{T}} + \frac{d}{T} + \frac{dG_{\infty}}{T^{1-s}} \bigg). 
% \end{align*}}
% When $\sum_{i=1}^d  \|\gb_{1:T,i}\|_2\ll \sqrt{dT}$ \citep{duchi2011adaptive,reddi2018convergence}, we further have
% {\small\begin{align}
%     \EE\Big[\big\|\nabla f(\xb_{\text{out}})\big\|_2^2\Big] = O\bigg(\frac{1}{\sqrt{T}} + \frac{d}{T} + \sqrt{\frac{d}{T}}\bigg),\notag
% \end{align}}
% which matches the convergence result in nonconvex SGD \citep{ghadimi2016accelerated} considering the dependence of $T$. %\CC{match the rate of SGD in Natasha2 paper?}
% \end{remark}

%Compared with their work, (i) our analysis applies directly to Padam, which is a more general algorithm comparing with those analyzed in \citet{chen2018on}; and (ii) our derived convergence rate is sharper than theirs.
%Another attempt to obtain the convergence rate of adaptive algorithms under stochastic nonconvex setting is prompted recently by 
% Need to cite several others. Ask me if you dont know which one

% \vspace{-0.2cm}
\section{Experiments}\label{sec:exp}
% \vspace{-0.1cm}
In this section, we empirically evaluate our proposed algorithm for training various modern deep learning models and test them on several standard benchmarks.\footnote{The code is available  
at \url{https://github.com/uclaml/Padam}.}
We show that for nonconvex loss functions in deep learning, our proposed algorithm still enjoys a fast convergence rate, while its generalization performance is as good as SGD with momentum and much better than existing adaptive gradient methods such as Adam and Amsgrad.  
% \subsection{Environmental Setup}
% All experiments are conducted on Amazon AWS p3.8xlarge servers which come with Intel Xeon E5 CPU and 4 NVIDIA Tesla V100 GPUs (16G RAM per GPU). 
% All experiments are implemented in Pytorch version $0.4.1$ within Python $3.6.4$.

\begin{figure*}[ht!]
    \centering
    \subfigure[Train Loss for VGGNet]{\includegraphics[width=0.32\textwidth]{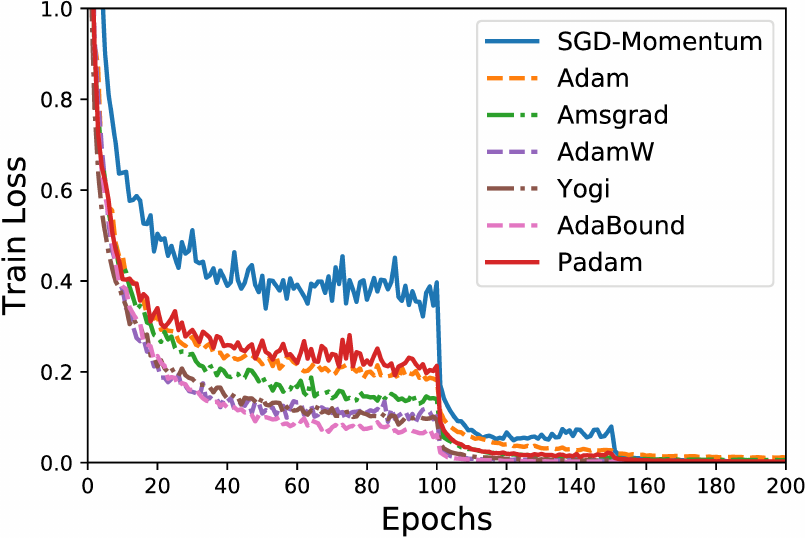}}
    \subfigure[Train Loss for ResNet]{\includegraphics[width=0.32\textwidth]{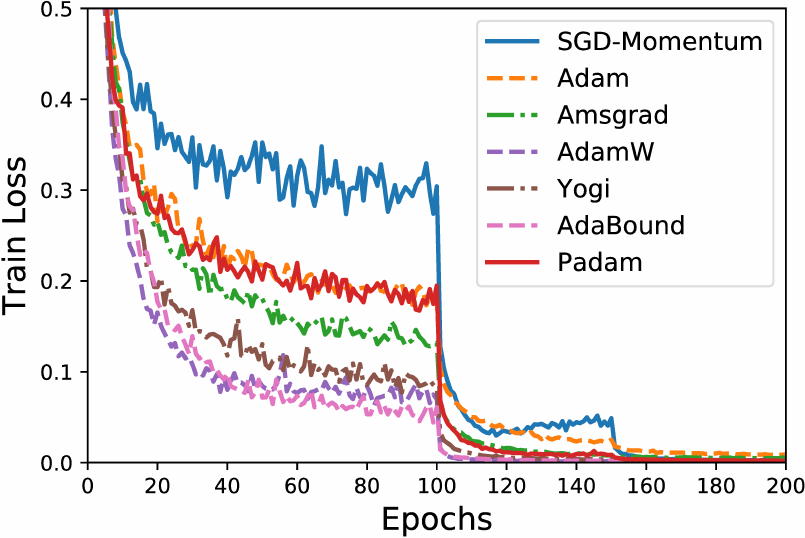}}
    \subfigure[Train Loss for WideResNet]{\includegraphics[width=0.32\textwidth]{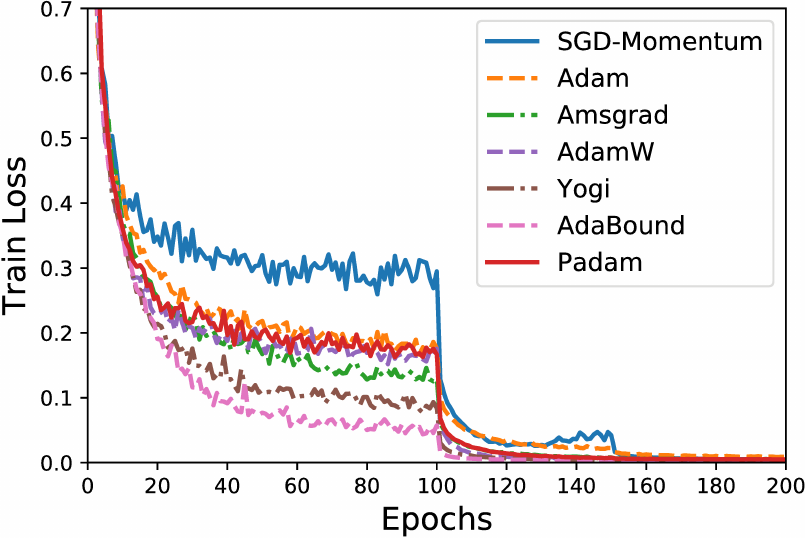}}
    \subfigure[Test Error for VGGNet]{\includegraphics[width=0.32\textwidth]{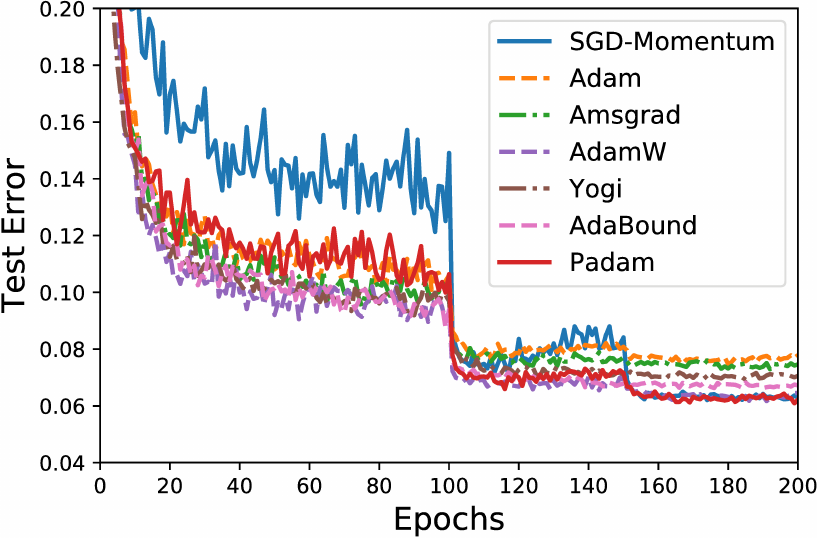}}
    \subfigure[Test Error for ResNet]{\includegraphics[width=0.32\textwidth]{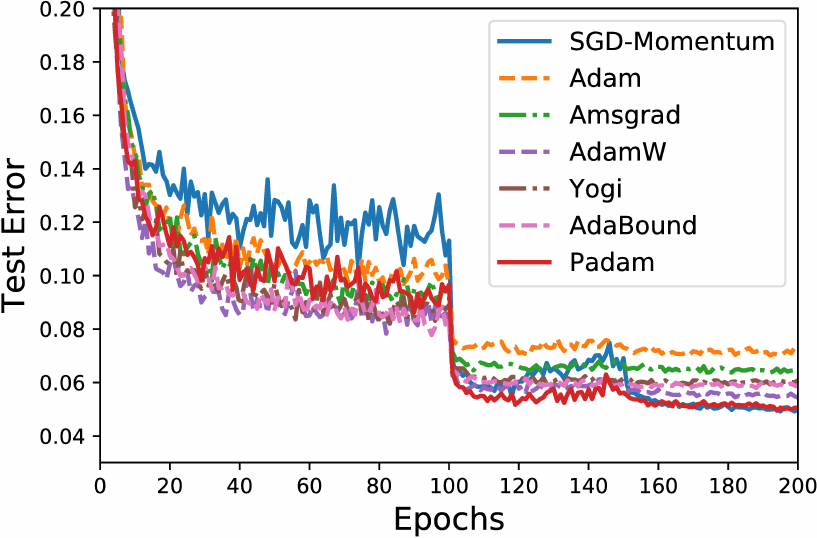}}
    \subfigure[Test Error for WideResNet]{\includegraphics[width=0.32\textwidth]{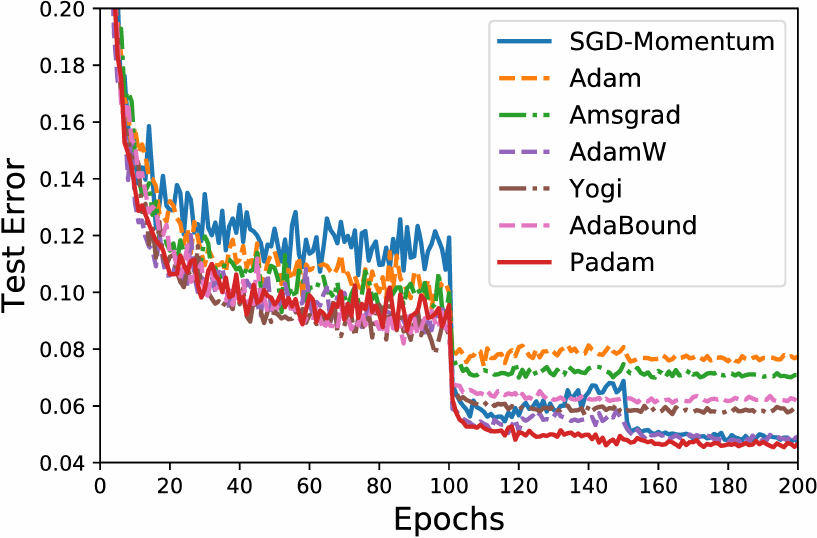}}
    \caption{Train loss and test error (top-$1$) on the CIFAR-10 dataset.}
    \label{fig:cifar10}
\end{figure*}

% \begin{figure}[ht!]
%     \centering
%     \subfigure[Train Loss for VGGNet]{\includegraphics[width=0.23\textwidth]{figure/cifar100_vggnet_trainloss.pdf}}
%     \subfigure[Train Loss for ResNet]{\includegraphics[width=0.23\textwidth]{figure/cifar100_resnet_trainloss.pdf}}
%     \subfigure[Train Loss for WideResNet]{\includegraphics[width=0.23\textwidth]{figure/cifar100_wideresnet_trainloss.pdf}}
%     \subfigure[Test Error for VGGNet]{\includegraphics[width=0.23\textwidth]{figure/cifar100_vggnet_testerr.pdf}}
%     \subfigure[Test Error for ResNet]{\includegraphics[width=0.23\textwidth]{figure/cifar100_resnet_testerr.pdf}}
%     \subfigure[Test Error for WideResNet]{\includegraphics[width=0.23\textwidth]{figure/cifar100_wideresnet_testerr.pdf}}
%     \caption{Train loss and test error (top-$1$) on the CIFAR-100 dataset.}
%     \label{fig:cifar100}
% \end{figure}

% \vspace{-0.3cm}
% \begin{wrapfigure}{r}{0.61\textwidth}
\begin{figure*}[ht!] 
% \vspace{-0.5cm}
    \centering
    \subfigure[Top-$1$ Error, VGGNet]{\includegraphics[width=0.32\textwidth]{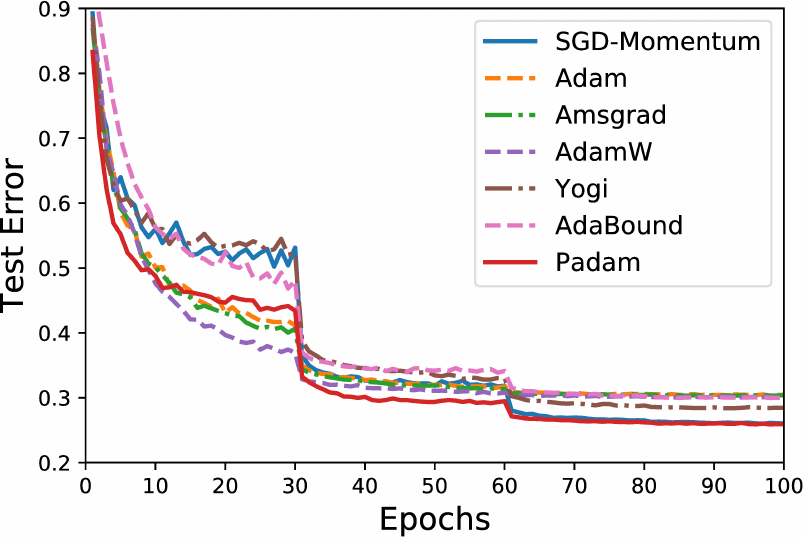}}
    \subfigure[Top-$1$ Error, ResNet]{\includegraphics[width=0.32\textwidth]{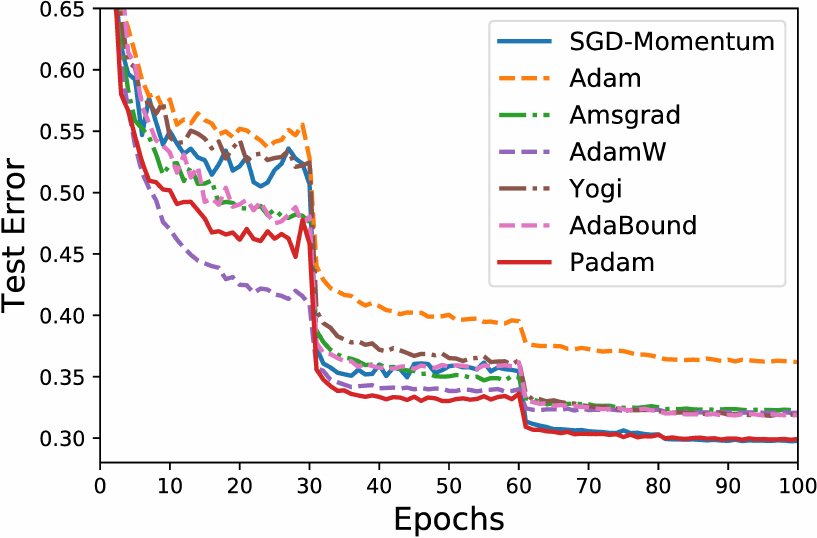}}
    \subfigure[2-layer LSTM]{\includegraphics[width=0.32\textwidth]{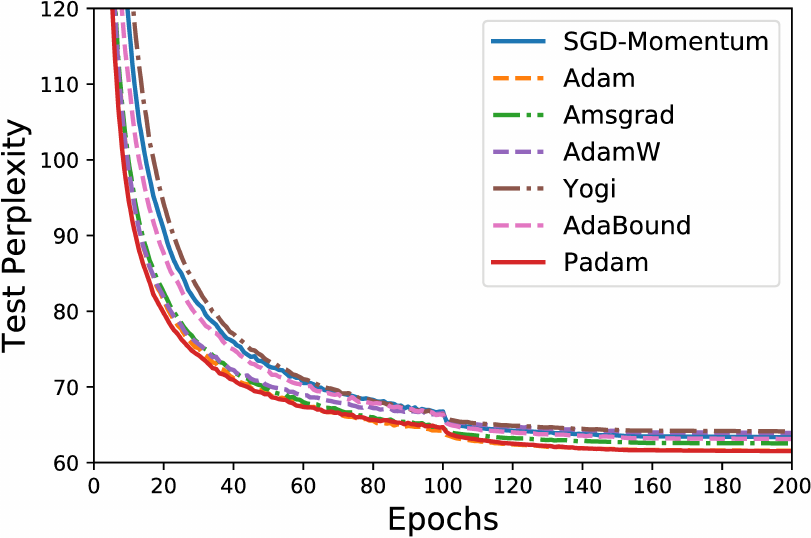}}
    \subfigure[Top-$5$ Error, VGGNet]{\includegraphics[width=0.32\textwidth]{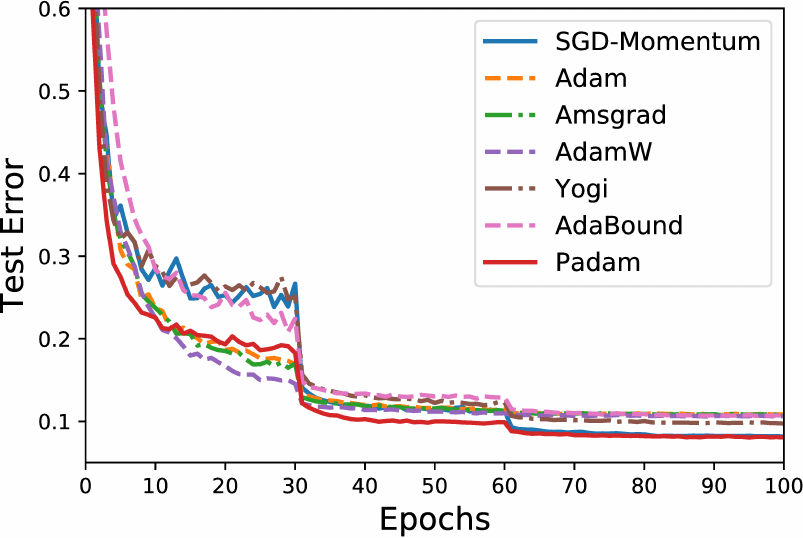}}
    \subfigure[Top-$5$ Error, ResNet]{\includegraphics[width=0.32\textwidth]{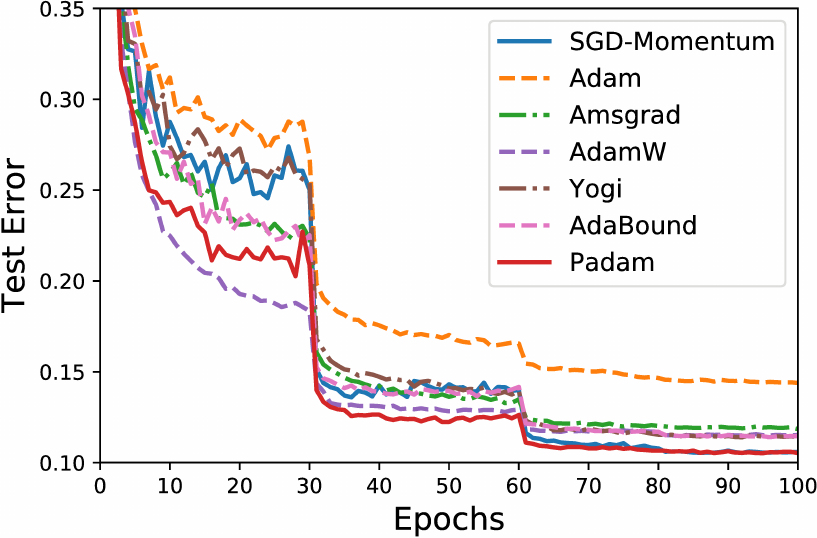}}
    \subfigure[3-layer  LSTM]{\includegraphics[width=0.32\textwidth]{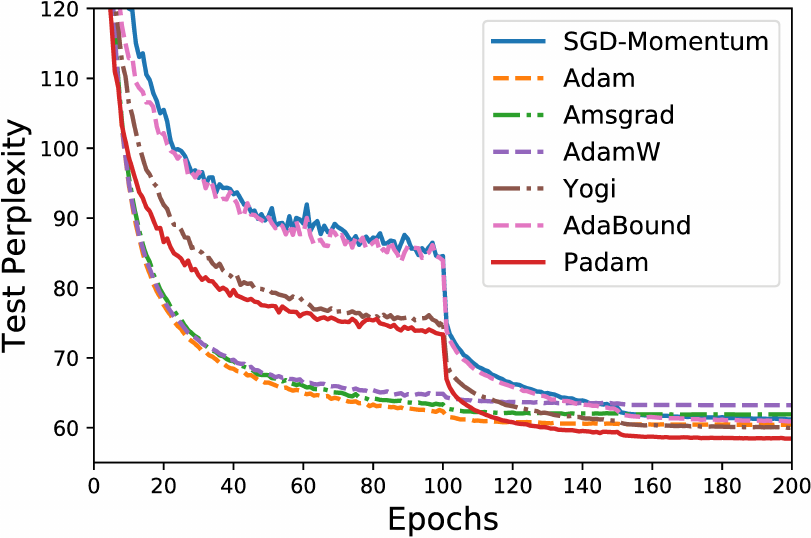}}
    \caption{Test error on the ImageNet dataset (left and middle columns), and test perplexity on the Penn Treebank dataset (right column).}
    \label{fig:imagenet-lstm}
\end{figure*} 
% \end{wrapfigure}

%  \begin{figure}[!t] 
%     \centering
%     \subfigure[2-layer LSTM]{\includegraphics[width=0.23\textwidth]{figure/LSTM_layer2_testppl.pdf}}
%     \subfigure[3-layer  LSTM]{\includegraphics[width=0.23\textwidth]{figure/LSTM_layer3_testppl.pdf}}
%     \caption{Test perplexity for 2-layer and 3-layer LSTM model on the Penn Treebank dataset.}
%     \label{fig:lstm}
% \end{figure} 

\begin{table*}[h]
  \begin{center}
  \begin{small}
%   \begin{sc}
  \begin{tabular}{c|ccccccc}
    \toprule
    % \multicolumn{1}{c}{\multirow{2}{*}{Models}} & \multicolumn{7}{c}{Test Accuracy (\%) }\\
    % % \cline{2-5}
    % \cmidrule(r){2-8}
    % \multicolumn{1}{c}{} & SGD-Momentum & Adam & Amsgrad  & AdamW & Yogi & AdaBound & Padam\\
    Models & SGD-Momentum & Adam & Amsgrad  & AdamW & Yogi & AdaBound & Padam\\

    \midrule
    VGGNet & $  93.71 $ & $  92.21 $ & $  92.54 $ & $  93.54 $ & $  92.94 $ & $  93.28 $ & $  \mathbf{93.78} $\\
    ResNet & $  \mathbf{95.00} $ & $  92.89 $ & $  93.53 $ & $  94.56 $ & $  93.92 $ & $  94.16 $ & $  94.94 $\\
    WideResNet & $  95.26 $ & $  92.27 $ & $  92.91 $ & $  95.08 $ & $  94.23 $ & $  93.85 $ & $  \mathbf{95.34} $\\
    \bottomrule
  \end{tabular}
%   \end{sc}
  \end{small}
  \end{center}
  \caption{Test accuracy (\%) of all algorithms after $200$ epochs on the CIFAR-10 dataset. Bold number indicates the best result.}  
  \label{table:cifar10}
\end{table*}

% \begin{table*}[ht!]
%   \begin{center}
%   \begin{small}
%   \begin{tabular}{c|ccccccc}
%     \toprule
%     \multicolumn{1}{c}{\multirow{2}{*}{Models}} & \multicolumn{7}{c}{Test Accuracy (\%) }\\
%     % \cline{2-5}
%     \cmidrule(r){2-8}
%     \multicolumn{1}{c}{} & SGDM & Adam & Amsgrad  & AdamW & Yogi & AdaBound & Padam\\
%     \midrule
%     VGGNet & $  73.32 $ & $  66.60 $ & $  69.40 $ & $  73.03 $ & $  72.35 $ & $  72.00 $ & $  \mathbf{74.39}$\\
%     ResNet & $  77.77 $ & $  71.72 $ & $  72.62 $ & $  76.69 $ & $  74.55 $ & $  75.29 $ & $  \mathbf{77.85} $\\
%     WideResNet & $  \mathbf{76.66} $ & $  71.83 $ & $  73.02 $ & $  76.04 $ & $  74.47 $ & $  73.49 $ & $  76.42 $ \\
%     \bottomrule
%   \end{tabular}
%   \end{small}
%   \end{center}
%   \caption{Test accuracy of all algorithms after $200$ epochs on the CIFAR-100 dataset. Bold number indicates the best result.}
%   \label{table:cifar100}
% \end{table*}

\begin{table*}[h]
    \centering
    \begin{small}
    \begin{tabular}{c|c|ccccccc}
    \toprule
    % \multicolumn{1}{c}{\multirow{2}{*}{Models}} & \multicolumn{7}{c}{Test Accuracy ($\%$)}\\
    % % \cline{2-5}
    % \cmidrule(r){2-8}
    % \multicolumn{1}{c}{} & SGD-Momentum & Adam & Amsgrad  & AdamW & Yogi & AdaBound & Padam\\
    Models & Test Accuracy & SGD-Momentum & Adam & Amsgrad  & AdamW & Yogi & AdaBound & Padam\\
         \midrule
         \multirow{2}{*}{Resnet} & Top-1  & $\mathbf{70.23}$ & $63.79$ & $67.69$ & $67.93$ & $68.23$ & $68.13$ & $70.07$\\
         & Top-5  & $89.40$ & $85.61$ & $ 88.15$ & $88.47$ & $88.59$ & $88.55$ & $\mathbf{89.47}$\\
         \midrule
         \multirow{2}{*}{VGGNet} & Top-1  & $73.93$ & $69.52$ & $69.61$ & $69.89$ & $71.56$ & $70.00$ & $\mathbf{74.04}$\\
         & Top-5  & $91.82$ & $89.12$ & $ 89.19$ & $89.35$ & $90.25$ & $89.27$ & $\mathbf{91.93}$\\
         \bottomrule
    \end{tabular}
    \end{small}
    \caption{Test accuracy (\%) of all algorithms after $100$ epochs on the ImageNet dataset. Bold number indicates the best result.}
    \label{tab:imagenet}
\end{table*}

\begin{table*}[h]
    \centering
    \begin{small}
    \begin{tabular}{c|ccccccc}
    \toprule
    Models & SGD-Momentum & Adam & Amsgrad  & AdamW & Yogi & AdaBound & Padam\\
         \midrule
         2-layer LSTM & $  63.37 $ & $  61.58 $ &$  62.56 $ &$  63.93 $ &$  64.13 $ &$  63.14 $ &$  \mathbf{61.53} $ \\
         3-layer LSTM  & $  61.22 $ &$  60.44 $ &$  61.92 $ &$  63.24 $ &$  60.01 $ &$  60.89 $ &$  \mathbf{58.48} $\\
         \bottomrule
    \end{tabular}
    \end{small}
    \caption{Test perplexity (lower is better) of all algorithms after $200$ epochs on the Penn Treebank dataset. Bold number indicates the best result.}
    \label{tab:lstm}
\end{table*}

We compare Padam against several state-of-the-art algorithms, including:
(1) SGD-Momentum, (2) Adam \citep{kingma2014adam}, (3) Amsgrad \citep{reddi2018convergence}, (4) AdamW \citep{loshchilov2017fixing} (5) Yogi \citep{manzil2018adaptive} and (6) AdaBound \citep{Luo2019AdaBound}.
We use several popular datasets for image classifications and language modeling: CIFAR-10 \citep{krizhevsky2009learning},
% CIFAR-100 \citep{krizhevsky2009learning}, 
ImageNet dataset (ILSVRC2012) \citep{deng2009imagenet} and Penn Treebank dataset \citep{marcus1993building}. 
We adopt three popular CNN architectures for image classification task: VGGNet-16 \citep{simonyan2014very}, Residual Neural Network (ResNet-18)  \citep{he2016deep}, Wide Residual Network (WRN-16-4) \citep{zagoruyko2016wide}. We test the language modeling task using 2-layer and 3-layer Long Short-Term Memory (LSTM) network \citep{hochreiter1997long}.
For CIFAR-10 and Penn Treebank experiments, 
we test for $200$ epochs and decay the learning rate by $0.1$ at the $100$th and $150$th epoch. We test ImageNet tasks for $100$ epochs with similar multi-stage learning rate decaying scheme at the $30$th, $60$th and $80$th epoch. 
% We perform grid search on validation set to choose the best hyper-parameters for each algorithm including learning rate, $\beta_1$ and $\beta_2$ and other algorithm-specific parameters such as  $p$ in Padam, $\epsilon$ in Yogi, weight decay for AdamW, final learning rate for AdaBound. Other parameters (batch size, etc.) are set to default value for each experimental setting.
% In terms of the choice of $p$, we recommend doing binary search from $1/4$ to $1/8$, to $1/16$, etc. Yet in most cases we tested, $1/8$ is a stable and reliable choice of $p$.
% Details about the datasets, CNN architectures and all the specific model parameters used in the following experiments can be found in the supplementary materials.

% \subsection{Parameter Settings}
We perform grid searches to choose the best hyper-parameters for all algorithms in both image classification and language modeling tasks.
For the base learning rate, we do grid search over $\{10^{-4}, \ldots,10^2\}$ for all algorithms, and choose the partial adaptive parameter $p$ from $\{2/5, 1/4, 1/5, 1/8, 1/16\}$ and the second order moment parameter $\beta_2$ from $\{0.9, 0.99, 0.999\}$.
For image classification experiments, we set
the base learning rate of $0.1$ for SGD with momentum and Padam, $0.001$ for all other adaptive gradient methods.
$\beta_1$ is set as $0.9$ for all methods. $\beta_2$ is set as $0.99$ for Adam and Amsgrad, $0.999$ for all other methods.
For Padam, the partially adaptive parameter $p$ is set to be $1/8$. For AdaBound, the final learning rate is set to be $0.1$. 
% The weight decay factor is set to $5\times 10^{-4}$ for all methods except AdamW which adopts a different weight decay scheme.
For AdamW, the normalized weight decay factor is set to $2.5\times 10^{-2}$ for CIFAR-10 and $5\times 10^{-2}$ for ImageNet. 
For Yogi, $\epsilon$ is set as $10^{-3}$ as suggested in the original paper. 
The minibatch size for CIFAR-10 is set to be $128$ and for ImageNet dataset we set it to be $256$. 
Regarding the LSTM experiments, for SGD with momentum, the base learning rate is set to be $1$ for 2-layer LSTM model and $10$ for 3-layer LSTM. The momentum parameter is set to be $0.9$ for both models. For all adaptive gradient methods except Padam and Yogi, we set the base learning rate as $0.001$. For Yogi, we set the base learning rate as $0.01$ for 2-layer LSTM model and $0.1$ for 3-layer LSTM model. For Padam, we set the base learning rate as $0.01$ for 2-layer LSTM model and $1$ for 3-layer LSTM model. For all adaptive gradient methods, we set $\beta_1 = 0.9$, $\beta_2 = 0.999$. In terms of algorithm specific parameters, for Padam, we set the partially adaptive parameter $p$ as $0.4$ for 2-layer LSTM model and $0.2$ for 3-layer LSTM model. For AdaBound, we set the final learning rate as $10$ for 2-layer LSTM model and $100$ for 3-layer LSTM model. For Yogi, $\epsilon$ is set as $10^{-3}$ as suggested in the original paper. 
% The weight decay factor is set to $1.2\times 10^{-6}$ for all methods except AdamW which adopts a different weight decay scheme.
For AdamW, the normalized weight decay factor is set to $4\times 10^{-4}$. 
The minibatch size is set to be $20$ for all LSTM experiments.

% \vspace{-0.2cm}
\subsection{Experimental Results}
We compare our proposed algorithm with other baselines on training the aforementioned three modern CNN architectures for image classification on the CIFAR-10 and ImageNet datasets. 
Figure \ref{fig:cifar10} plots the train loss and test error (top-$1$ error) against training epochs on the CIFAR-10 dataset. As we can see from the figure, at the early stage of the training process, (partially) adaptive gradient methods including Padam, make rapid progress lowing both the train loss and the test error, while SGD with momentum converges relatively slowly. After the first learning rate decaying at the $100$-th epoch, different algorithms start to behave differently. SGD with momentum makes a huge drop while fully adaptive gradient methods (Adam and Amsgrad) start to generalize badly. Padam, on the other hand, maintains relatively good generalization performance and also holds the lead over other algorithms. After the second decaying at the $150$-th epoch, Adam and Amsgrad basically lose all the power of traversing through the parameter space due to the ``small learning dilemma'', while the performance of SGD with momentum finally catches up with Padam. AdamW, Yogi and AdaBound indeed improve the performance compared with original Adam but the performance is still worse than Padam. 
% \cc{
% Similar performances can be observed in Figure \ref{fig:cifar100}, where we plot the train loss and test error (top-$1$ error) against training epochs on the CIFAR-100 dataset.
% }
Overall we can see that Padam achieves the best of both worlds (i.e., Adam and SGD with momentum): it maintains faster convergence rate while also generalizing as well as SGD with momentum in the end.

Figure \ref{fig:imagenet-lstm} (a)(b)(d)(e) plot the Top-$1$ and Top-$5$ error against training epochs on the ImageNet dataset for both VGGNet and ResNet. We can see that on the ImageNet dataset, all methods behave similarly as in our CIFAR-10 experiments. Padam method again obtains the best from both worlds by achieving the fastest convergence while generalizing as well as SGD with momentum. Even though methods such as AdamW, Yogi and AdaBound have better performance than standard Adam, they still suffer from a big generalization gap on the ImageNet dataset.
Note that we did not conduct WideResNet experiment on the Imagenet dataset due to GPU memory limits.

% We also test the influence of different partial adaptive parameter $p$ on the performance of our proposed method. 
% Figure \ref{fig:cifar10_dist}(a) shows the comparison of test error performances under the different partial adaptive parameter $p$ for ResNet on CIFAR-10 dataset. 
% We can observe that a larger $p$ will lead to fast convergence at early stages and worse generalization performance later, while a smaller $p$ behaves more like SGD with momentum: slow in early stages but finally catch up. While a proper choice of $p$ (e.g., $1/8$), could obtain the best of both worlds.  

We also perform experiments on the language modeling tasks to test our proposed algorithm on Long Short-Term Memory (LSTM) network \citep{hochreiter1997long}, where adaptive gradient methods such as Adam are currently the mainstream optimizers for these tasks. 
Figure \ref{fig:imagenet-lstm} (c)(f) plot the test perplexity against training epochs on the Penn Treebank dataset \citep{marcus1993building} for both 2-layer LSTM and 3-layer LSTM models. We can observe that the differences on simpler 2-layer LSTM model is not very obvious but on more complicated 3-layer LSTM model, different algorithms have quite different optimizing behaviors. Even though Adam, Amsgrad and AdamW have faster convergence in the early stages, Padam achieves the best final test perplexity on this language modeling task for both of our experiments.

% \cc{
For a more quantitative comparison, we also provide the test accuracy for all above experiments.
Table \ref{table:cifar10} shows the test accuracy of all algorithms on the CIFAR-10 dataset.
On the CIFAR-10 dataset, methods such as Adam and Amsgrad have the lowest test accuracy. Even though more recent algorithms AdamW, Yogi, AdaBound improve upon original Adam, they still fall behind or barely match the performance of SGD with momentum. In contrast, Padam achieves the highest test accuracy for VGGNet and WideResNet on CIFAR-10 dataset. For training ResNet on the CIFAR-10 dataset, Padam is also on a par with SGD with momentum at the final epoch (differences less than $0.2\%$).
% }
% \cc{
Table \ref{tab:imagenet} shows the final test accuracy of all algorithms on the ImageNet dataset. Again, we can observe that Padam achieves the best test accuracy for VGGNet (both Top-1 and Top-5) and Top-1 accuracy for ResNet. It stays very close to the best baseline of Top-1 accuracy for the ResNet model.
% } 
% \cc{
Table \ref{tab:lstm} shows the final test perplexity of all algorithms on the Penn Treebank dataset. As we can see, Padam achieves the best (lowest) test perplexity on both 2-layer LSTM and 3-layer LSTM models.
% } 
All these experimental results suggest that practitioners can use Padam for training deep neural networks, without worrying about the generalization performances.

% \vspace{-0.2cm}
\section{Conclusions and Future Work}\label{sec:conclusion}
% \section{Conclusions}\label{sec:conclusion}
% \vspace{-0.2cm}
In this paper, we proposed Padam, which unifies Adam/Amsgrad with SGD-Momentum. With an appropriate choice of the partially adaptive parameter, we show that Padam can achieve the best from both worlds, i.e., maintaining fast convergence rate while closing the generalization gap. We also provide a theoretical analysis towards the convergence rate of Padam to a stationary point for stochastic nonconvex optimization.

% While the empirical generalization performances achieved by Padam backup our hypothesis of ``small learning rate dilemma", it is still unclear  how learning rate decay affects the performance of adaptive gradient methods in theory. We leave it as an important future direction. 

%\CC{the first future work has been done} 
% On the other hand, while the regret bound delivered in this paper is informative, it is restricted to convex functions as in previous analyses of Adam \citep{kingma2014adam} and Amsgrad \citep{reddi2018convergence}. An important and challenging future direction is to analyze the convergence of adaptive gradient methods including our algorithm for nonconvex functions. Since our algorithm can also be seen as an interpolation between Amsgrad and SGD with momentum, we can borrow some idea from the convergence analysis of SGD with momentum \citep{kidambi2018insufficiency} to facilitate the analysis in this direction. 
 
It would also be interesting to see how well Padam performs in other types of neural networks, such as generative adversarial network (GAN) \citep{goodfellow2014generative} and graph convolutional neural network (GCN) \citep{kipf2016semi,zou2019layer}. We leave it as a future work.

\section*{Acknowledgements}
%\cc{This research was sponsored in part by the National Science
%Foundation XXX.}

We thank the anonymous reviewers for their helpful comments. This research was sponsored in part by the National Science Foundation CAREER Award IIS-1906169, BIGDATA IIS-1855099 and IIS-1903202. We also thank AWS for providing cloud computing credits associated with the NSF BIGDATA
award. The views and conclusions contained in this paper are those of the authors and should not be interpreted as representing any funding agencies.
 
\bibliography{padam}
\bibliographystyle{named}

\appendix
\newpage
\onecolumn
\section{Proof of the Main Theory}

In this section, we provide a detailed version of proof of Theorem \ref{eq:Gu0000}.
\subsection{Proof of Theorem \ref{eq:Gu0000}}
Let $\xb_0 = \xb_1$ and 
\begin{align}
    \zb_{t} = \xb_{t} + \frac{\beta_1}{1-\beta_1}(\xb_{t} - \xb_{t-1}) = \frac{1}{1-\beta_1}\xb_{t} - \frac{\beta_1}{1 - \beta_1}\xb_{t-1}, \label{def:z}
\end{align}
we have the following lemmas:

\begin{lemma}\label{lm:3}
Let $\zb_t$ be defined in \eqref{def:z}. For $t \geq 2$, we have
\begin{align}
\zb_{t+1}-\zb_{t}&= \frac{\beta_1}{1 - \beta_1}\Big[\Ib - \big(\alpha_t\hat{\Vb}_{t}^{-p}\big) \big(\alpha_{t-1}\hat{\Vb}_{t-1}^{-p}\big)^{-1}\Big](\xb_{t-1} - \xb_t) - \alpha_t\hat{\Vb}_{t}^{-p} \gb_t. \label{lm:7.3.1}
\end{align}
and 
\begin{align}
   \zb_{t+1}-\zb_{t} &  = 
\frac{\beta_1}{1 - \beta_1}\big(\alpha_{t-1}\hat{\Vb}_{t-1}^{-p} - \alpha_t\hat{\Vb}_{t}^{-p}\big)\mb_{t-1} - \alpha_t \hat{\Vb}_{t}^{-p} \gb_t.\label{lm:7.3.2}
\end{align}
For $t = 1$, we have
\begin{align}
    \zb_2 - \zb_1 = -\alpha_1 \hat{\Vb}_{1}^{-p}\gb_1.
\end{align}
\end{lemma}

\begin{lemma}\label{lm:4}
Let $\zb_t$ be defined in \eqref{def:z}. For $t \geq 2$, we have
\begin{align*}
\|\zb_{t+1}-\zb_{t}\|_2&\le\big\|\alpha\hat{\Vb}_t^{-p}\gb_t\big\|_2+\frac{\beta_1}{1 - \beta_1}\|\xb_{t-1}-\xb_{t}\|_2.
\end{align*}
\end{lemma}

\begin{lemma}\label{lm:5}
Let $\zb_t$ be defined in \eqref{def:z}. For $t \geq 2$, we have
\begin{align*}
\|\nabla f(\zb_t)-\nabla f(\xb_t)\|_2&\leq L\Big(\frac{\beta_1}{1 - \beta_1}\Big)\cdot\|\xb_t-\xb_{t-1}\|_2.
\end{align*}
\end{lemma}

\begin{lemma}\label{lm:1}
Let $\hat{\vb}_t$ and $\mb_t$ be as defined in Algorithm \ref{alg:Padam}. Then under Assumption~\ref{as:1}, we have $\|\nabla f(\xb)\|_\infty \leq G_\infty$, $\|\hat{\vb}_t\|_\infty \leq G_\infty^2$ and $\|\mb_{t}\|_\infty \leq G_\infty$.
\end{lemma}

% \begin{lemma}\label{lm:11.1}

%  For Algorithm \ref{algo3}, when $p\in[0,1/2]$, $q\in(\max\{0,4p-1\},1]$, we have\\
%  \begin{align*}
%  \sum_{t=1}^{T}\alpha_t\EE\big[\|\Hat{\Vb}_t^{-p}\mb_t\|_2^2\big] &\leq \frac{\alpha G_\infty^{(1+q-4p)}T^{q/2}}{(1-\beta_1)(1-\gamma)(1-\beta_2)^{2p}}\sum_{i=1}^{d}\|\gb_{1:T,i}\|_{2}^{1-q},
%  \end{align*}
%  where we define $\gamma:=\beta_1/\beta_2^{2p}$.
%  Specially, when $p\in[0,1/4]$, $q=0$, we get
%   \begin{align*}
%  \sum_{t=1}^{T}\alpha_t\EE\big[\|\Hat{\Vb}_t^{-p}\mb_t\|_2^2\big] &\leq \frac{\alpha G_\infty^{(1-4p)}\sqrt{1+\log T}}{(1-\beta_1)(1-\gamma)(1-\beta_2)^{2p}}\sum_{i=1}^{d}\|\gb_{1:T,i}\|_{2}.
%  \end{align*}
%  \end{lemma}

\begin{lemma}\label{lm:2}
Suppose that $f$ has $G_\infty$-bounded stochastic gradient. Let $\beta_1, \beta_2$ be the weight parameters, $\alpha_t$, $t=1,\ldots,T$ be the step sizes in Algorithm \ref{alg:Padam} and $q \in [\max\{4p-1,0\}, 1]$. We denote $\gamma = \beta_1 / \beta_2^{2p}$. Suppose that $\alpha_t = \alpha$ and $\gamma \leq 1$, then under Assumption~\ref{as:1}, we have the following two results:
\begin{align*}
     \sum_{t=1}^T \alpha_t^2\EE\Big[\big\|\Hat{\Vb}_t^{-p}\mb_t\big\|_2^2\Big]
     & \leq \frac{T^{(1+q)/2}d^q\alpha^2(1-\beta_1) G_\infty^{(1+q-4p)}}{(1-\beta_2)^{2p}(1-\gamma)} \EE \bigg(\sum_{i=1}^d \|\gb_{1:T,i}\|_2\bigg)^{1-q},
 \end{align*}
and
 \begin{align*}
     \sum_{t=1}^T \alpha_t^2\EE\Big[\big\|\Hat{\Vb}_t^{-p}\gb_t\big\|_2^2\Big]
     & \leq \frac{T^{(1+q)/2}d^q\alpha^2 G_\infty^{(1+q-4p)}}{(1-\beta_2)^{2p}} \EE \bigg(\sum_{i=1}^d \|\gb_{1:T,i}\|_2\bigg)^{1-q}.
 \end{align*}
\end{lemma}
%  \begin{lemma}\label{lm:7.2}

% For Algorithm \ref{algo3}, and $p>0$ we have\\
% \begin{align*}
% \|\oneb-\hat{\vb}_t^{-p}\|^2_{\hat{\Vb}_t^{2p}}\le\sqrt{d}(G_\infty^{4p}+1).
% \end{align*}
% \end{lemma}
%To deal with stochastic momentum $\mb_t$ and stochastic weight $\hat{\Vb}_t^{-p}$, following \citet{yang2016unified}, we define an auxiliary sequence $\zb_t$ as follows: let $\xb_0 = \xb_1$, and for each $t \geq 1$, 

%Lemma \ref{lm:3} shows that $\zb_{t+1} - \zb_t$ can be represented in two different ways.

%By Lemma \ref{lm:3}, we connect $\zb_{t+1}-\zb_{t}$ with $\xb_{t+1}-\xb_t$ and $\alpha_t\Hat{\Vb}_t^{-p}\gb_t$

%The following two lemmas give bounds on $\|\zb_{t+1} - \zb_t\|_2$ and $\|\nabla f(\zb_t)-\nabla f(\xb_t)\|_2$, which play important roles in our proof.

Now we are ready to prove Theorem \ref{eq:Gu0000}.

\begin{proof}[Proof of Theorem \ref{eq:Gu0000}]
Since $f$ is $L$-smooth, we have:
\begin{align}
f(\zb_{t+1})&\le  f(\zb_t)+\nabla f(\zb_t)^\top(\zb_{t+1}-\zb_t)+\frac{L}{2}\|\zb_{t+1}-\zb_t\|_2^2\notag \\
& =
f(\zb_t) +\underbrace{\nabla f(\xb_t)^\top(\zb_{t+1}-\zb_t)}_{I_1}+\underbrace{(\nabla f(\zb_t) -\nabla f(\xb_t))^\top(\zb_{t+1}-\zb_t)}_{I_2} +\underbrace{\frac{L}{2}\|\zb_{t+1}-\zb_t\|_2^2}_{I_3}\label{newtheorem_0} 
\end{align}
In the following, we bound $I_1$, $I_2$ and $I_3$ separately. 

\noindent\textbf{Bounding term $I_1$:} When $t = 1$, we have
\begin{align}
    \nabla f(\xb_1)^\top(\zb_{2}-\zb_1) = -\nabla f(\xb_1)^\top\alpha_{1} \hat{\Vb}_{t}^{-p} \gb_1.\label{newtheorem_4.5}
\end{align}
For $t \geq 2$, we have
\begin{align}
    &\nabla f(\xb_t)^\top(\zb_{t+1}-\zb_t) \notag \\
    & =
    \nabla f(\xb_t)^\top\bigg[\frac{\beta_1}{1 - \beta_1}\big(\alpha_{t-1}\hat{\Vb}_{t-1}^{-p} - \alpha_t\hat{\Vb}_{t}^{-p}\big)\mb_{t-1} - \alpha_t \hat{\Vb}_{t}^{-p} \gb_t\bigg]\notag \\
    & = 
    \frac{\beta_1}{1-\beta_1}\nabla f(\xb_t)^\top\big(\alpha_{t-1}\hat{\Vb}_{t-1}^{-p} - \alpha_t\hat{\Vb}_{t}^{-p}\big)\mb_{t-1} - \nabla f(\xb_t)^\top\alpha_t \hat{\Vb}_{t}^{-p} \gb_t,\label{newtheorem_1}
\end{align}
where the first equality holds due to \eqref{lm:7.3.2} in Lemma \ref{lm:3}. For $\nabla f(\xb_t)^\top(\alpha_{t-1}\hat{\Vb}_{t-1}^{-p} - \alpha_t\hat{\Vb}_{t}^{-p})\mb_{t-1}$ in \eqref{newtheorem_1}, we have
\begin{align}
    \nabla f(\xb_t)^\top(\alpha_{t-1}\hat{\Vb}_{t-1}^{-p} - \alpha_t\hat{\Vb}_{t}^{-p})\mb_{t-1} &\leq \|\nabla f(\xb_t)\|_\infty\cdot \big\|\alpha_{t-1}\hat{\Vb}_{t-1}^{-p} - \alpha_t\hat{\Vb}_{t}^{-p}\big\|_{1,1}\cdot \|\mb_{t-1}\|_\infty \notag \\
    & \leq 
    G_\infty^2 \Big[\big\|\alpha_{t-1}\hat{\Vb}_{t-1}^{-p}\big\|_{1,1} - \big\|\alpha_t\hat{\Vb}_{t}^{-p}\big\|_{1,1}\Big]\notag \\
    & = 
    G_\infty^2 \Big[\big\|\alpha_{t-1}\hat{\vb}_{t-1}^{-p}\big\|_1 - \big\|\alpha_t\hat{\vb}_{t}^{-p}\big\|_1\Big].\label{newtheorem_2}
\end{align}
The first inequality holds because for a positive diagonal matrix $\Ab$, we have $\xb^\top\Ab\yb\leq \|\xb\|_\infty\cdot\|\Ab\|_{1,1}\cdot\|\yb\|_\infty$. The second inequality holds due to $\alpha_{t-1} \hat{\Vb}_{t-1}^{-p} \succeq \alpha_{t} \hat{\Vb}_{t}^{-p}\succeq 0$. Next we bound $-\nabla f(\xb_t)^\top\alpha_t \hat{\Vb}_{t}^{-p} \gb_t$. We have
\begin{align}
    &-\nabla f(\xb_t)^\top\alpha_t \hat{\Vb}_{t}^{-p} \gb_t \notag \\
    & = 
    -\nabla f(\xb_t)^\top\alpha_{t-1} \hat{\Vb}_{t-1}^{-p} \gb_t - \nabla f(\xb_t)^\top\big(\alpha_t \hat{\Vb}_{t}^{-p} -\alpha_{t-1} \hat{\Vb}_{t-1}^{-p} \big) \gb_t\notag \\
    & \leq 
    -\nabla f(\xb_t)^\top\alpha_{t-1} \hat{\Vb}_{t-1}^{-p} \gb_t + \| \nabla f(\xb_t)\|_\infty \cdot \big\|\alpha_t \hat{\Vb}_{t}^{-p} -\alpha_{t-1} \hat{\Vb}_{t-1}^{-p}\big\|_{1,1}  \cdot  \|\gb_t\|_\infty\notag \\
    & \leq 
    -\nabla f(\xb_t)^\top\alpha_{t-1} \hat{\Vb}_{t-1}^{-p} \gb_t+ G_\infty^2 \Big(\big\|\alpha_{t-1} \hat{\Vb}_{t-1}^{-p}\big\|_{1,1} - \big\|\alpha_{t} \hat{\Vb}_{t}^{-p}\big\|_{1,1}\Big)\notag\\
    & = 
    -\nabla f(\xb_t)^\top\alpha_{t-1} \hat{\Vb}_{t-1}^{-p} \gb_t+ G_\infty^2 \Big(\big\|\alpha_{t-1} \hat{\vb}_{t-1}^{-p}\big\|_1 - \big\|\alpha_{t} \hat{\vb}_{t}^{-p}\big\|_1\Big).\label{newtheorem_3}
\end{align}
The first inequality holds because for a positive diagonal matrix $\Ab$, we have $\xb^\top\Ab\yb\leq \|\xb\|_\infty\cdot\|\Ab\|_{1,1}\cdot\|\yb\|_\infty$. The second inequality holds due to $\alpha_{t-1} \hat{\Vb}_{t-1}^{-p} \succeq \alpha_{t} \hat{\Vb}_{t}^{-p}\succeq 0$. Substituting \eqref{newtheorem_2} and \eqref{newtheorem_3} into \eqref{newtheorem_1}, we have
\begin{align}
    &\nabla f(\xb_t)^\top(\zb_{t+1}-\zb_t)  \leq 
    -\nabla f(\xb_t)^\top\alpha_{t-1} \hat{\Vb}_{t-1}^{-p} \gb_t + \frac{1}{1-\beta_1}G_\infty^2 \Big(\big\|\alpha_{t-1} \hat{\vb}_{t-1}^{-p}\big\|_1 - \big\|\alpha_{t} \hat{\vb}_{t}^{-p}\big\|_1\Big).\label{newtheorem_4}
\end{align}
%Next we bound $(\nabla f(\zb_t) -\nabla f(\xb_t))^\top(\zb_{t+1}-\zb_t)$. 
\noindent\textbf{Bounding term $I_2$:} For $t\geq 1$, we have
\begin{align}
    &\big(\nabla f(\zb_t) -\nabla f(\xb_t)\big)^\top(\zb_{t+1}-\zb_t) \notag \\
    & \leq 
    \big\|\nabla f(\zb_t) -\nabla f(\xb_t)\big\|_2\cdot\|\zb_{t+1}-\zb_t\|_2\notag \\
    & \leq 
    \Big(\big\|\alpha_t\hat{\Vb}_t^{-p}\gb_t\big\|_2+\frac{\beta_1}{1 - \beta_1}\|\xb_{t-1}-\xb_{t}\|_2\Big)\cdot \frac{\beta_1}{1 - \beta_1}\cdot L\|\xb_t-\xb_{t-1}\|_2\notag\\
    & = 
    L\frac{\beta_1}{1 - \beta_1}\big\|\alpha_t\hat{\Vb}_t^{-p}\gb_t\big\|_2\cdot\|\xb_t-\xb_{t-1}\|_2 + L \bigg(\frac{\beta_1}{1 - \beta_1}\bigg)^2\|\xb_t-\xb_{t-1}\|_2^2\notag \\
    & \leq 
    L\big\|\alpha_t\hat{\Vb}_t^{-p}\gb_t\big\|_2^2 + 2L \bigg(\frac{\beta_1}{1 - \beta_1}\bigg)^2\|\xb_t-\xb_{t-1}\|_2^2,\label{newtheorem_5}
\end{align}
where the second inequality holds because of Lemma \ref{lm:3} and Lemma \ref{lm:4}, the last inequality holds due to Young's inequality. 

\noindent\textbf{Bounding term $I_3$:} %Now we bound $\frac{L}{2}\cdot \|\zb_{t_1} - \zb_t\|_2^2$. 
For $t\geq 1$, we have
\begin{align}
    \frac{L}{2}\|\zb_{t+1} - \zb_t\|_2^2 
    & \leq 
    \frac{L}{2}\Big[\big\|\alpha_t\hat{\Vb}_t^{-p}\gb_t\big\|_2+\frac{\beta_1}{1 - \beta_1}\|\xb_{t-1}-\xb_{t}\|_2\Big]^2\notag \\
    & \leq 
    L\big\|\alpha_t\hat{\Vb}_t^{-p}\gb_t\big\|_2^2+2L\bigg(\frac{\beta_1}{1 - \beta_1}\bigg)^2\|\xb_{t-1}-\xb_{t}\|_2^2.\label{newtheorem_6}
\end{align}
The first inequality is obtained by introducing  Lemma \ref{lm:3}. 

For $t = 1$, substituting \eqref{newtheorem_4.5}, \eqref{newtheorem_5} and \eqref{newtheorem_6} into \eqref{newtheorem_0}, taking expectation and rearranging terms,
we have
\begin{align}
    &\EE[f(\zb_2) - f(\zb_1)] \notag \\
    &\leq \EE\bigg[-\nabla f(\xb_1)^\top\alpha_{1} \hat{\Vb}_{1}^{-p} \gb_1+2L\big\|\alpha_1\hat{\Vb}_1^{-p}\gb_1\big\|_2^2 + 4L \bigg(\frac{\beta_1}{1 - \beta_1}\bigg)^2\|\xb_1-\xb_{0}\|_2^2\bigg]\notag \\
    & = 
    \EE[-\nabla f(\xb_1)^\top\alpha_{1} \hat{\Vb}_{1}^{-p} \gb_1+2L\big\|\alpha_1\hat{\Vb}_1^{-p}\gb_1\big\|_2^2 ]\notag \\
    & \leq
    \EE[d \alpha_1 G_\infty^{2-2p} + 2L\big\|\alpha_1\hat{\Vb}_1^{-p}\gb_1\big\|_2^2], \label{newtheorem_7.5}
\end{align}
where the last inequality holds because 
\begin{align}
    -\nabla f(\xb_1)^\top\hat{\Vb}_{1}^{-p} \gb_1 \leq d\cdot \|\nabla f(\xb_1)\|_\infty\cdot \|\hat{\Vb}_1^{-p}\gb_1\|_\infty \leq d\cdot G_\infty\cdot G_\infty^{1-2p} = dG_\infty^{2-2p}.\notag
\end{align}
For $t\geq 2$, substituting \eqref{newtheorem_4}, \eqref{newtheorem_5} and \eqref{newtheorem_6} into \eqref{newtheorem_0}, taking expectation and rearranging terms, we have
\begin{align}
    &\EE\bigg[f(\zb_{t+1})+\frac{G_\infty^2\big\|\alpha_{t} \hat{\vb}_{t}^{-p}\big\|_1}{1-\beta_1} - \bigg(f(\zb_t) + \frac{G_\infty^2\big\|\alpha_{t-1} \hat{\vb}_{t-1}^{-p}\big\|_1}{1-\beta_1}\bigg)\bigg] \notag \\
    &\leq 
    \EE\bigg[-\nabla f(\xb_t)^\top\alpha_{t-1} \hat{\Vb}_{t-1}^{-p} \gb_t +  2L\big\|\alpha_t\hat{\Vb}_t^{-p}\gb_t\big\|_2^2 + 4L \bigg(\frac{\beta_1}{1 - \beta_1}\bigg)^2\|\xb_t-\xb_{t-1}\|_2^2\bigg]\notag \\
    & =
    \EE\bigg[-\nabla f(\xb_t)^\top\alpha_{t-1} \hat{\Vb}_{t-1}^{-p} \nabla f(\xb_t) +  2L\big\|\alpha_t\hat{\Vb}_t^{-p}\gb_t\big\|_2^2 + 4L \bigg(\frac{\beta_1}{1 - \beta_1}\bigg)^2\big\|\alpha_{t-1}\hat{\Vb}_{t-1}^{-p}\mb_{t-1}\big\|_2^2\bigg]\notag \\
    & \leq 
    \EE\bigg[-\alpha_{t-1}\big\|\nabla f(\xb_t)\big\|_2^2(G_\infty^{2p})^{-1} +  2L\big\|\alpha_t\hat{\Vb}_t^{-p}\gb_t\big\|_2^2 + 4L \bigg(\frac{\beta_1}{1 - \beta_1}\bigg)^2\big\|\alpha_{t-1}\hat{\Vb}_{t-1}^{-p}\mb_{t-1}\big\|_2^2\bigg],\label{newtheorem_7}
\end{align}
where the equality holds because $\EE [\gb_t] = \nabla f(\xb_t)$ conditioned on $\nabla f(\xb_t)$ and $\hat{\Vb}_{t-1}^{-p}$, the second inequality holds because of Lemma \ref{lm:1}. 
Telescoping \eqref{newtheorem_7} for $t=2$ to $T$ and adding with \eqref{newtheorem_7.5}, we have
\begin{align}
    &(G_\infty^{2p})^{-1}\sum_{t=2}^T \alpha_{t-1}\EE\big\|\nabla f(\xb_t)\big\|_2^2 \notag \\
    &\leq 
    \EE\bigg[f(\zb_1) + \frac{G_\infty^2\big\|\alpha_{1} \hat{\vb}_{1}^{-p}\big\|_1}{1-\beta_1} + d\alpha_1G_\infty^{2-2p} - \bigg(f(\zb_{T+1})+\frac{G_\infty^2\big\|\alpha_{T} \hat{\vb}_{T}^{-p}\big\|_1}{1-\beta_1}\bigg)\bigg]\notag \\
    &\quad\quad + 
    2L \sum_{t=1}^T\EE \big\|\alpha_t\hat{\Vb}_t^{-p}\gb_t\big\|_2^2 + 4L \bigg(\frac{\beta_1}{1 - \beta_1}\bigg)^2 \sum_{t=2}^T\EE\Big[ \big\|\alpha_{t-1}\hat{\Vb}_{t-1}^{-p}\mb_{t-1}\big\|_2^2\Big]\notag \\
    & \leq 
    \EE\bigg[\Delta f + \frac{G_\infty^2\big\|\alpha_{1} \hat{\vb}_{1}^{-p}\big\|_1}{1-\beta_1}+d\alpha_1G_\infty^{2-2p}\bigg]+ 
    2L \sum_{t=1}^T\EE \big\|\alpha_t\hat{\Vb}_t^{-p}\gb_t\big\|_2^2\nonumber\\
    &\qquad+ 4L \bigg(\frac{\beta_1}{1 - \beta_1}\bigg)^2 \sum_{t=1}^T\EE\Big[ \big\|\alpha_{t}\hat{\Vb}_{t}^{-p}\mb_{t}\big\|_2^2\Big].\label{newtheorem_8}
\end{align}
By Lemma \ref{lm:2}, we have
\begin{align}
     \sum_{t=1}^T \alpha_t^2\EE\Big[\|\Hat{\Vb}_t^{-p}\mb_t\|_2^2\Big]
     & \leq \frac{T^{(1+q)/2}d^q\alpha^2(1-\beta_1) G_\infty^{(1+q-4p)}}{(1-\beta_2)^{2p}(1-\gamma)} \EE \bigg(\sum_{i=1}^d \|\gb_{1:T,i}\|_2\bigg)^{1-q},\label{newtheorem_9}
 \end{align}
 where $\gamma = \beta_1/\beta_2^{2p}$. We also have 
 \begin{align}
     \sum_{t=1}^T \alpha_t^2\EE\Big[\|\Hat{\Vb}_t^{-p}\gb_t\|_2^2\Big]
     & \leq \frac{T^{(1+q)/2}d^q\alpha^2 G_\infty^{(1+q-4p)}}{(1-\beta_2)^{2p}} \EE \bigg(\sum_{i=1}^d \|\gb_{1:T,i}\|_2\bigg)^{1-q}.\label{newtheorem_10}
 \end{align}
Substituting \eqref{newtheorem_9} and \eqref{newtheorem_10} into \eqref{newtheorem_8}, and rearranging \eqref{newtheorem_8}, we have
\begin{align}
    &\EE\|\nabla f(\xb_{\text{out}})\|_2^2 \notag \\
    &= \frac{1}{\sum_{t=2}^T \alpha_{t-1}}\sum_{t=2}^T \alpha_{t-1}\EE\big\|\nabla f(\xb_t)\big\|_2^2 \notag \\
    & \leq 
    \frac{G_\infty^{2p}}{\sum_{t=2}^T \alpha_{t-1}}\EE\bigg[\Delta f+ \frac{G_\infty^2\big\|\alpha_{1} \hat{\vb}_{1}^{-p}\big\|_1}{1-\beta_1}+d\alpha_1G_\infty^{2-2p}\bigg]\notag \\
    &\quad\quad + 
    \frac{2LG_\infty^{2p}}{\sum_{t=2}^T \alpha_{t-1}}\frac{T^{(1+q)/2}d^q\alpha^2 G_\infty^{(1+q-4p)}}{(1-\beta_2)^{2p}} \EE \bigg(\sum_{i=1}^d \|\gb_{1:T,i}\|_2\bigg)^{1-q}\notag \\
    &\quad\quad + 
    \frac{4LG_\infty^{2p}}{\sum_{t=2}^T \alpha_{t-1}}\bigg(\frac{\beta_1}{1 - \beta_1}\bigg)^2\frac{T^{(1+q)/2}d^q\alpha^2(1-\beta_1) G_\infty^{(1+q-4p)}}{(1-\beta_2)^{2p}(1-\gamma)} \EE \bigg(\sum_{i=1}^d \|\gb_{1:T,i}\|_2\bigg)^{1-q}\notag \\
    & \leq 
    \frac{1}{T\alpha}2G_\infty^{2p}\Delta f + \frac{4}{T}\bigg(\frac{G_\infty^{2+2p}\EE\big\| \hat{\vb}_{1}^{-p}\big\|_1}{1-\beta_1}+dG_\infty^{2}\bigg)\notag \\
    &\quad\quad + 
   \frac{d^{q}\alpha}{T^{(1-q)/2}}\EE \bigg(\sum_{i=1}^d \|\gb_{1:T,i}\|_2\bigg)^{1-q} \bigg(\frac{4LG_\infty^{1+q-2p}}{(1-\beta_2)^{2p}} + \frac{8LG_\infty^{1+q-2p}(1-\beta_1)}{(1-\beta_2)^{2p}(1-\gamma)}\bigg(\frac{\beta_1}{1 - \beta_1}\bigg)^2\bigg),\label{newtheorem_11}
\end{align}
where the second inequality holds because $\alpha_t = \alpha$. Rearranging \eqref{newtheorem_11}, we obtain
\iffalse
\begin{align}
    \EE\|\nabla f\|_2^2 \leq \frac{1}{T\alpha}+\frac{d}{T}+\alpha d^q\bigg(\frac{\sum}{T^{1/2}}\bigg)^{1-q}
\end{align}
\fi
\begin{align}
    \EE\|\nabla f(\xb_{\text{out}})\|_2^2 \leq \frac{M_1}{T\alpha} + \frac{M_2 d}{T}+\frac{\alpha d^qM_3}{T^{(1-q)/2}}\EE \bigg(\sum_{i=1}^d  \|\gb_{1:T,i}\|_2\bigg)^{1-q},\notag
\end{align}
where $\{M_i\}_{i=1}^3$ are defined in Theorem \ref{eq:Gu0000}.
% \begin{align*}
%     M_1 &= 2G_\infty^{2p}\Delta f,\quad
%     M_2 = \frac{4G_\infty^{2+2p}\EE\big\| \hat{\vb}_{1}^{-p}\big\|_1/d}{1-\beta_1} + 4G_\infty^2, \\
%     M_3 &= \frac{4LG_\infty^{1+q-2p}}{(1-\beta_2)^{2p}} + \frac{8LG_\infty^{1+q-2p}(1-\beta_1)}{(1-\beta_2)^{2p}(1-\gamma)}\bigg(\frac{\beta_1}{1 - \beta_1}\bigg)^2,
% \end{align*}
This completes the proof. 
\end{proof}

\subsection{Proof of Corollary \ref{cl:1}}
\begin{proof}[Proof of Corollary \ref{cl:1}]
From Theorem \ref{eq:Gu0000}, let $p\in[0,1/4]$, we have $q\in[0,1]$.
Setting $q=0$, we have
\begin{align}
    \EE\Big[\big\|\nabla f(\xb_{\text{out}})\big\|_2^2\Big] \leq \frac{M_1}{T\alpha} + \frac{M_2\cdot d}{T}+ \frac{M_3'\alpha}{\sqrt{T}}\EE \bigg(\sum_{i=1}^d  \|\gb_{1:T,i}\|_2\bigg),\notag
\end{align}
where $M_1$ and $M_2$ are defined in Theorem \ref{eq:Gu0000} and $M_3'$ is defined in Corollary \ref{cl:1}. This completes the proof.
% \begin{align}
%     M_3'  = \frac{4LG_\infty^{1-2p}}{(1-\beta_2)^{2p}} + \frac{8LG_\infty^{1-2p}(1-\beta_1)}{(1-\beta_2)^{2p}(1-\beta_1/\beta_2^{2p})}\bigg(\frac{\beta_1}{1 - \beta_1}\bigg)^2.\notag
% \end{align}
\end{proof}

\section{Proof of Technical Lemmas}

\subsection{Proof of Lemma \ref{lm:3}}
\begin{proof}%[Proof of Lemma \ref{lm:3}]
By definition, we have
\begin{align}
    \zb_{t+1} = \xb_{t+1} + \frac{\beta_1}{1 - \beta_1}(\xb_{t+1} - \xb_{t}) = \frac{1}{1 - \beta_1}\xb_{t+1} - \frac{\beta_1}{1 - \beta_1}\xb_{t}.\notag 
\end{align}
Then we have
\begin{align}
    \zb_{t+1} - \zb_t &= \frac{1}{1 - \beta_1}(\xb_{t+1} - \xb_t) - \frac{\beta_1}{1 - \beta_1}(\xb_t - \xb_{t-1}) \notag \\
    & = 
    \frac{1}{1 - \beta_1}\big(-\alpha_t\hat{\Vb}_t^{-p}\mb_t\big) + \frac{\beta_1}{1 - \beta_1}\alpha_{t-1}\hat{\Vb}_{t-1}^{-p}\mb_{t-1} \notag.
\end{align}
The equities above are based on definition. Then we have
\begin{align}
    \zb_{t+1} - \zb_t & = \frac{-\alpha_t\hat{\Vb}_t^{-p} }{1 - \beta_1}\Big[\beta_1\mb_{t-1}+(1 - \beta_1) \gb_t\Big] + \frac{\beta_1}{1 - \beta_1}\alpha_{t-1}\hat{\Vb}_{t-1}^{-p}\mb_{t-1} \notag \\
    & = 
    \frac{\beta_1}{1 - \beta_1}\mb_{t-1}\big(\alpha_{t-1}\hat{\Vb}_{t-1}^{-p} - \alpha_t\hat{\Vb}_{t}^{-p}\big) - \alpha_t \hat{\Vb}_{t}^{-p} \gb_t\notag \\
    & = 
    \frac{\beta_1}{1 - \beta_1}\alpha_{t-1}\hat{\Vb}_{t-1}^{-p}\mb_{t-1}\Big[\Ib - \big(\alpha_t\hat{\Vb}_{t}^{-p}\big) \big(\alpha_{t-1}\hat{\Vb}_{t-1}^{-p}\big)^{-1}\Big] - \alpha_t\hat{\Vb}_{t}^{-p} \gb_t\notag \\
    & = 
    \frac{\beta_1}{1 - \beta_1}\Big[\Ib - \big(\alpha_t\hat{\Vb}_{t}^{-p}\big) \big(\alpha_{t-1}\hat{\Vb}_{t-1}^{-p}\big)^{-1}\Big](\xb_{t-1} - \xb_t) - \alpha_t\hat{\Vb}_{t}^{-p} \gb_t\notag.
\end{align}
The equalities above follow by combining the like terms. %.
\end{proof}

\subsection{Proof of Lemma \ref{lm:4}}
\begin{proof}%[Proof of Lemma \ref{lm:4}]
By Lemma \ref{lm:3}, we have
\begin{align*}
\|\zb_{t+1}-\zb_{t}\|_2&=\bigg\|\frac{\beta_1}{1 - \beta_1}\Big[\Ib - (\alpha_t\hat{\Vb}_{t}^{-p}) (\alpha_{t-1}\hat{\Vb}_{t-1}^{-p})^{-1}\Big](\xb_{t-1} - \xb_t) - \alpha_t\hat{\Vb}_{t}^{-p} \gb_t\bigg\|_2\\
&\le\frac{\beta_1}{1 - \beta_1}\Big\|\Ib - (\alpha_t\hat{\Vb}_{t}^{-p}) (\alpha_{t-1}\hat{\Vb}_{t-1}^{-p})^{-1}\Big\|_{\infty,\infty}\cdot\|\xb_{t-1}-\xb_t\|_2+\big\|\alpha\hat{\Vb}_t^{-p}\gb_t\big\|_2,
\end{align*}
where the inequality holds because the term $\beta_1/(1 - \beta_1)$ is positive, and triangle inequality. Considering that $\alpha_t\hat{\vb}_{t,j}^{-p}\le\alpha_{t-1}\hat{\vb}_{t-1,j}^{-p}$, when $p>0$, we have $\Big\|\Ib - (\alpha_t\hat{\Vb}_{t}^{-p}) (\alpha_{t-1}\hat{\Vb}_{t-1}^{-p})^{-1}\Big\|_{\infty, \infty}\le 1$. With that fact, the term above can be bound as:
\begin{align*}
\|\zb_{t+1}-\zb_{t}\|_2&\le\big\|\alpha\hat{\Vb}_t^{-p}\gb_t\big\|_2+\frac{\beta_1}{1 - \beta_1}\|\xb_{t-1}-\xb_{t}\|_2.
\end{align*}
This completes the proof.
\end{proof}

\subsection{Proof of Lemma \ref{lm:5}}
\begin{proof}%[Proof of Lemma \ref{lm:5}]
For term $\|\nabla f(\zb_t)-\nabla f(\xb_t)\|_2$, we have:
\begin{align*}
\|\nabla f(\zb_t)-\nabla f(\xb_t)\|_2&\le L\|\zb_t-\xb_t\|_2\\
&\leq L\Big\|\frac{\beta_1}{1 - \beta_1}(\xb_{t} - \xb_{t-1})\Big\|_2\\
&\leq L\Big(\frac{\beta_1}{1 - \beta_1}\Big)\cdot\|\xb_t-\xb_{t-1}\|_2,
\end{align*}
where the last inequality holds because the term $\beta_1/(1 - \beta_1)$ is positive.
\end{proof}

\iffalse
\begin{align}
\sum_{i=1}^d\sqrt{\log T+1}\cdot\sqrt{\sum_{t=1}^T\frac{|g_{t,i}|^2}{t}}
\end{align}
\begin{align}
&\sqrt{\sum_{t=1}^T\frac{|g_{t,i}|^2}{t}}\\
=&\sqrt{\sum_{t=1}^{\sqrt{T}}\frac{|g_{t,i}|^2}{t}+\sum_{t=\sqrt{T}}^{T}\frac{|g_{t,i}|^2}{t}}\\
=&\sqrt{\sum_{t=1}^{\sqrt{T}}|g_{t,i}|^2+\frac{1}{\sqrt{T}}\sum_{t=\sqrt{T}}^{T}|g_{t,i}|^2}\\
=&\sqrt{\sum_{t=1}^{\sqrt{T}}|g_{t,i}|^2}+\sqrt{\frac{1}{\sqrt{T}}\sum_{t=\sqrt{T}}^{T}|g_{t,i}|^2}
=\|\gb_{1:\sqrt{T}},i\|_2+\frac{1}{\sqrt[4]{T}}\|\gb_{1:T,i}\|_2
\end{align}
\fi

\subsection{Proof of Lemma \ref{lm:1}}
\begin{proof}[Proof of Lemma \ref{lm:1}]
Since $f$ has $G_\infty$-bounded stochastic gradient, for any $\xb$ and $\xi$, $\|\nabla f(\xb; \xi)\|_\infty \leq G_\infty$. Thus, we have 
\begin{align}
    \|\nabla f(\xb)\|_\infty = \|\EE_\xi\nabla f(\xb;\xi)\|_\infty \leq \EE_\xi\|\nabla f(\xb;\xi)\|_\infty \leq G_\infty. \notag
\end{align}
Next we bound $\|\mb_t\|_\infty$. We have $\|\mb_0\| = 0 \leq G_\infty$. Suppose that $\|\mb_t\|_\infty \leq G_\infty$, then for $\mb_{t+1}$, we have
\begin{align*}
    \|\mb_{t+1}\|_\infty &=  \|\beta_1\mb_{t}+(1-\beta_1)\gb_{t+1}\|_\infty\\ 
    &\leq \beta_1\|\mb_t\|_\infty  + (1-\beta_1)\|\gb_{t+1}\|_\infty \\
    &\leq \beta_1G_\infty + (1-\beta_1) G_\infty \\
    &= G_\infty.\notag
\end{align*}
Thus, for any $t \geq 0$, we have $\|\mb_t\|_\infty \leq G_\infty$. Finally we bound $\|\hat{\vb}_t\|_\infty$. First we have $\|\vb_0\|_\infty = \|\hat{\vb}_0\|_\infty=0\leq G_\infty^2$. Suppose that $\|\hat{\vb}_t\|_\infty \leq G_\infty^2$ and $\|\vb_t\|_\infty \leq G_\infty^2$. Note that we have
\begin{align*}
     \|\vb_{t+1}\|_\infty&= \|\beta_2\vb_t+(1-\beta_2)\gb_{t+1}^2\|_\infty \\
    &\leq \beta_2\|\vb_t\|_\infty+(1-\beta_2)\|\gb_{t+1}^2\|_\infty \\
    &\leq \beta_2 G_\infty^2+(1-\beta_2)G_\infty^2\\
    &=G_\infty^2,\notag
\end{align*}
and by definition, we have $\|\hat\vb_{t+1}\|_\infty = \max\{\|\hat\vb_t\|_\infty, \|\vb_{t+1}\|_\infty\} \leq G_\infty^2$. 
Thus, for any $t \geq 0$, we have $\|\hat{\vb}_t\|_\infty \leq G_\infty^2$.
\end{proof}

\subsection{Proof of Lemma \ref{lm:2}}

\begin{proof}%[Proof of Lemma \ref{lm:2}]
 Recall that $\hat{v}_{t,j}, m_{t,j}, g_{t,j}$ denote the $j$-th coordinate of $\hat{\vb}_t, \mb_t$ and $\gb_t$. We have
 \begin{align}
 \alpha_t^2\EE\Big[\|\Hat{\Vb}_t^{-p}\mb_t\|_2^2\Big]&
 =\alpha_t^2\EE\bigg[\sum_{i=1}^{d}\frac{m_{t,i}^2}{\hat{v}_{t,i}^{2p}}\bigg]\nonumber\\
 &\leq \alpha_t^2\EE\bigg[\sum_{i=1}^{d}\frac{m_{t,i}^2}{v_{t,i}^{2p}}\bigg]\nonumber\\
 &= \alpha_t^2\EE\bigg[\sum_{i=1}^{d}\frac{(\sum_{j=1}^t(1-\beta_1)\beta_1^{t-j}g_{j,i})^2}{(\sum_{j=1}^t(1-\beta_2)\beta_2^{t-j}g_{j,i}^2)^{2p}}\bigg], \label{lm:11.1_0}
 \end{align}
 where the first inequality holds because $\hat{v}_{t,i} \geq v_{t,i}$. Next we have 
  \begin{align}
 &\alpha_t^2\EE\bigg[\sum_{i=1}^{d}\frac{(\sum_{j=1}^t(1-\beta_1)\beta_1^{t-j}g_{j,i})^2}{(\sum_{j=1}^t(1-\beta_2)\beta_2^{t-j}g_{j,i}^2)^{2p}}\bigg]\notag\\
 &\leq \frac{\alpha_t^2(1-\beta_1)^2}{(1-\beta_2)^{2p}}\EE\bigg[\sum_{i=1}^{d}\frac{(\sum_{j=1}^t\beta_1^{t-j}|g_{j,i}|^{(1+q-4p)})(\sum_{j=1}^t\beta_1^{t-j}|g_{j,i}|^{(1-q+4p)})}{(\sum_{j=1}^t\beta_2^{t-j}g_{j,i}^2)^{2p}}\bigg]\notag\\
 &\leq \frac{\alpha_t^2(1-\beta_1)^2}{(1-\beta_2)^{2p}}\EE\bigg[\sum_{i=1}^{d}\frac{(\sum_{j=1}^t\beta_1^{t-j}G_\infty^{(1+q-4p)})(\sum_{j=1}^t\beta_1^{t-j}|g_{j,i}|^{(1-q+4p)})}{(\sum_{j=1}^t\beta_2^{t-j}g_{j,i}^2)^{2p}}\bigg]\notag\\
 &\leq \frac{\alpha_t^2(1-\beta_1) G_\infty^{(1+q-4p)}}{(1-\beta_2)^{2p}}\EE\bigg[\sum_{i=1}^{d}\frac{\sum_{j=1}^t\beta_1^{t-j}|g_{j,i}|^{(1-q+4p)}}{(\sum_{j=1}^t\beta_2^{t-j}g_{j,i}^2)^{2p}}\bigg],\label{lm:11.1_1}
\end{align}
where the first inequality holds due to Cauchy inequality, the second inequality holds because $|g_{j,i}| \leq G_\infty$, the last inequality holds because $\sum_{j=1}^t \beta_1^{t-j} \leq (1-\beta_1)^{-1}$. Note that
\begin{align}
 \sum_{i=1}^{d}\frac{\sum_{j=1}^t\beta_1^{t-j}|g_{j,i}|^{(1-q+4p)}}{(\sum_{j=1}^t\beta_2^{t-j}g_{j,i}^2)^{2p}}
 &\leq \sum_{i=1}^{d}\sum_{j=1}^{t}\frac{\beta_1^{t-j}|g_{j,i}|^{(1-q+4p)}}{(\beta_2^{t-j}g_{j,i}^2)^{2p}}= \sum_{i=1}^d\sum_{j=1}^{t}\gamma^{t-j}|g_{j,i}|^{1-q},\label{lm:11.1_2}
 \end{align}
 where the equality holds due to the definition of $\gamma$. Substituting \eqref{lm:11.1_1} and \eqref{lm:11.1_2} into \eqref{lm:11.1_0}, we have
 \begin{align}
     \alpha_t^2\EE\Big[\|\Hat{\Vb}_t^{-p}\mb_t\|_2^2\Big] \leq \frac{\alpha_t^2(1-\beta_1) G_\infty^{(1+q-4p)}}{(1-\beta_2)^{2p}} \EE \bigg[\sum_{i=1}^d\sum_{j=1}^{t}\gamma^{t-j}|g_{j,i}|^{1-q}\bigg].\label{lm:11.1_3}
 \end{align}
 Telescoping \eqref{lm:11.1_3} for $t=1$ to $T$, we have
 \begin{align}
     \sum_{t=1}^T \alpha_t^2\EE\Big[\|\Hat{\Vb}_t^{-p}\mb_t\|_2^2\Big]
     & \leq 
     \frac{\alpha^2(1-\beta_1) G_\infty^{(1+q-4p)}}{(1-\beta_2)^{2p}} \EE \bigg[\sum_{t=1}^T\sum_{i=1}^d\sum_{j=1}^{t}\gamma^{t-j}|g_{j,i}|^{1-q}\bigg] \notag \\
     & =
     \frac{\alpha^2(1-\beta_1) G_\infty^{(1+q-4p)}}{(1-\beta_2)^{2p}} \EE \bigg[\sum_{i=1}^d\sum_{j=1}^T|g_{j,i}|^{1-q}\sum_{t=j}^{T}\gamma^{t-j}\bigg] \notag \\
     & \leq 
     \frac{\alpha^2(1-\beta_1) G_\infty^{(1+q-4p)}}{(1-\beta_2)^{2p}(1-\gamma)} \EE \bigg[\sum_{i=1}^d\sum_{j=1}^T|g_{j,i}|^{1-q}\bigg]. \label{lm:11.1_4}
 \end{align}
 Finally, we have
 \begin{align}
     \sum_{i=1}^d\sum_{j=1}^T|g_{j,i}|^{1-q} 
     & \leq \sum_{i=1}^d \Big(\sum_{j=1}^T g_{j,i}^2\Big)^{(1-q)/2}\cdot T^{(1+q)/2} \notag \\
     & = 
     T^{(1+q)/2}\sum_{i=1}^d\|\gb_{1:T,i}\|_2^{1-q} \notag \\
     & \leq 
     T^{(1+q)/2}d^q \bigg(\sum_{i=1}^d \|\gb_{1:T,i}\|_2\bigg)^{1-q}, \label{lm:11.1_5}
 \end{align}
 where the first and second inequalities hold due to H\"{o}lder's inequality. Substituting \eqref{lm:11.1_5} into \eqref{lm:11.1_4}, we have
 \begin{align}
     \sum_{t=1}^T \alpha_t^2\EE\Big[\|\Hat{\Vb}_t^{-p}\mb_t\|_2^2\Big]
     & \leq \frac{T^{(1+q)/2}d^q\alpha^2(1-\beta_1) G_\infty^{(1+q-4p)}}{(1-\beta_2)^{2p}(1-\gamma)} \EE \bigg(\sum_{i=1}^d \|\gb_{1:T,i}\|_2\bigg)^{1-q}.\notag
 \end{align}
 Specifically, taking $\beta_1 = 0$, we have $\mb_t = \gb_t$, then
 \begin{align}
     \sum_{t=1}^T \alpha_t^2\EE\Big[\|\Hat{\Vb}_t^{-p}\gb_t\|_2^2\Big]
     & \leq \frac{T^{(1+q)/2}d^q\alpha^2 G_\infty^{(1+q-4p)}}{(1-\beta_2)^{2p}} \EE \bigg(\sum_{i=1}^d \|\gb_{1:T,i}\|_2\bigg)^{1-q}.\notag
 \end{align}
\end{proof}

\section{Experiment Details}

\subsection{Datasets}
We use several popular datasets for image classifications.
\begin{itemize}
% [leftmargin=*]
    \item CIFAR-10 \citep{krizhevsky2009learning}: it consists of a training set of $50,000$ $32 \times 32$ color images from $10$ classes, and also $10,000$ test images. 
    \item CIFAR-100 \citep{krizhevsky2009learning}: it is similar to CIFAR-10 but contains $100$ image classes with $600$ images for each. \item ImageNet dataset (ILSVRC2012)  \citep{deng2009imagenet}: ILSVRC2012 contains $1.28$ million training images, and $50k$ validation images over $1000$ classes. 
\end{itemize}
In addition, we adopt Penn Treebank (PTB) dataset \citep{marcus1993building}, which is widely used in Natural Language Processing (NLP) research. 
Note that word-level PTB dataset does not contain capital letters, numbers, and punctuation.
Models are evaluated based on the perplexity metric (lower is better).

\subsection{Architectures}
\noindent\textbf{VGGNet} \citep{simonyan2014very}: We use a modified VGG-16 architecture for this experiment. The VGG-16 network uses only $3 \times 3$ convolutional layers stacked on top of each other for increasing depth and adopts max pooling to reduce volume size. Finally, two fully-connected layers~\footnote{For CIFAR experiments, we change the ending two fully-connected layers from $2048$ nodes to $512$ nodes. For ImageNet experiments, we use batch normalized version (vgg16\_bn) provided in Pytorch official package} are followed by a softmax classifier. 
% In total it has approximately $26$ million parameters.

\noindent\textbf{ResNet} \citep{he2016deep}: Residual Neural Network (ResNet) introduces a novel architecture with ``skip connections'' and features a heavy use of batch normalization. Such skip connections are also known as gated units or gated recurrent units and have a strong similarity to recent successful elements applied in RNNs. We use ResNet-18 for this experiment, which contains $2$ blocks for each type of basic convolutional building blocks in \citet{he2016deep}. 
% In total it has approximately $1.8$ million parameters. 

\noindent\textbf{Wide ResNet} \citep{zagoruyko2016wide}: Wide Residual Network  further exploits the ``skip connections'' used in ResNet and in the meanwhile increases the width of residual networks. In detail, we use the $16$ layer Wide ResNet with $4$ multipliers (WRN-16-4) in our experiments.
% , which contains approximately $11$ million parameters.

\noindent\textbf{LSTM} \citep{hochreiter1997long}:
Long Short-Term Memory (LSTM) network is a special kind of Recurrent Neural Network (RNN), capable of learning long-term dependencies. It was introduced by \cite{hochreiter1997long}, and was refined and popularized in many followup work.

\section{Additional Experiments}

% \begin{figure}[t]
%   \centering
%     \includegraphics[width=0.4\textwidth]{figure/v_min_max.pdf}
%     \caption{Plot of max and min values across all coordinates of $\hat\vb_t$ against the iteration number for ResNet model on CIFAR10 dataset.}
%     \label{fig:max_min_v}
% \end{figure}

% As a sanity check, we first plot the max and min of $\hat{\vb}_t$ across all the coordinates in Figure \ref{fig:max_min_v}. In detail, the maximum value of $\hat{\vb}_t$ is around $0.04$ in the end and the minimum value is around $3\times10^{-8}$, $\max\{(\hat \vb_t)_i^{1/8}\} - \min\{(\hat \vb_t)_i^{1/8}\}\approx 0.55$. We can see that the effective learning rates for different coordinates are quite different.

% As a sanity check, we first plot the infinity norm of ${\hat \vb_t}$ for different network architectures in Figure \ref{fig:maxv}. As we can see, the infinity norm of ${\hat \vb_t}$ for most commonly used network architectures is strictly less than $1$ and therefore the adaptive term $1/\sqrt{\hat \vb_t}$ will indeed cause explosion with large learning rate.

\begin{figure*}[ht]
    \centering
    \subfigure[Train Loss for VGGNet]{\includegraphics[width=0.32\textwidth]{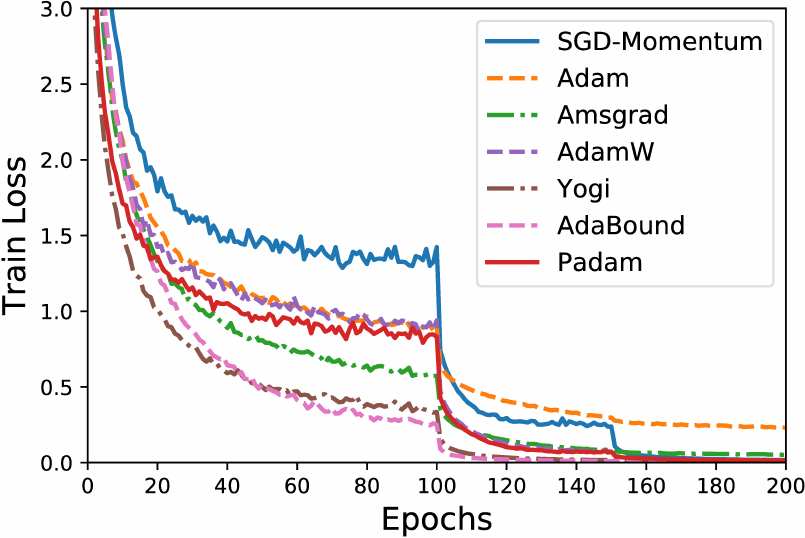}}
    \subfigure[Train Loss for ResNet]{\includegraphics[width=0.32\textwidth]{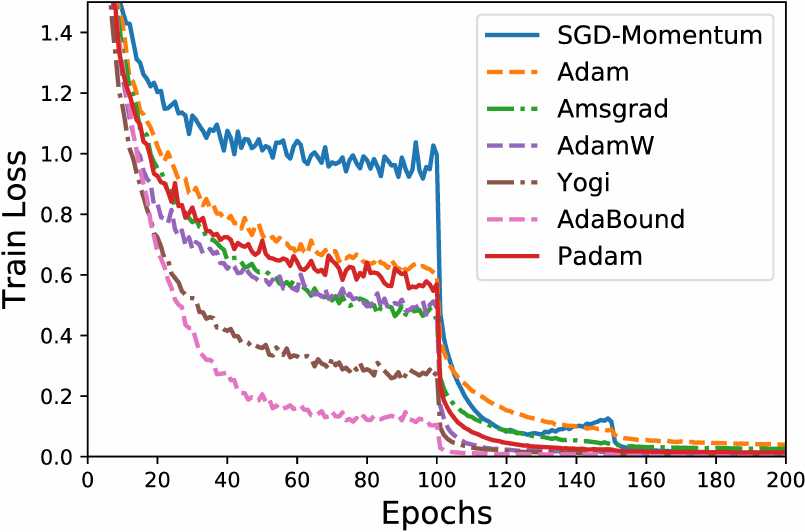}}
    \subfigure[Train Loss for WideResNet]{\includegraphics[width=0.32\textwidth]{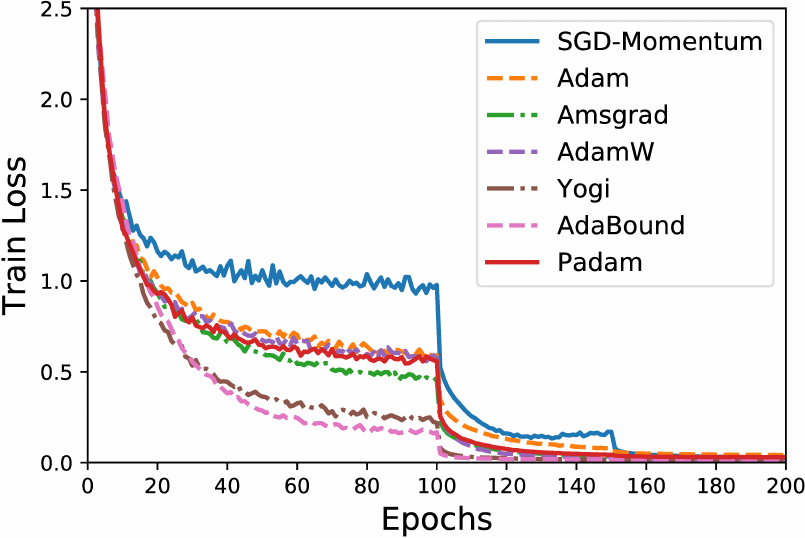}}
    \subfigure[Test Error for VGGNet]{\includegraphics[width=0.32\textwidth]{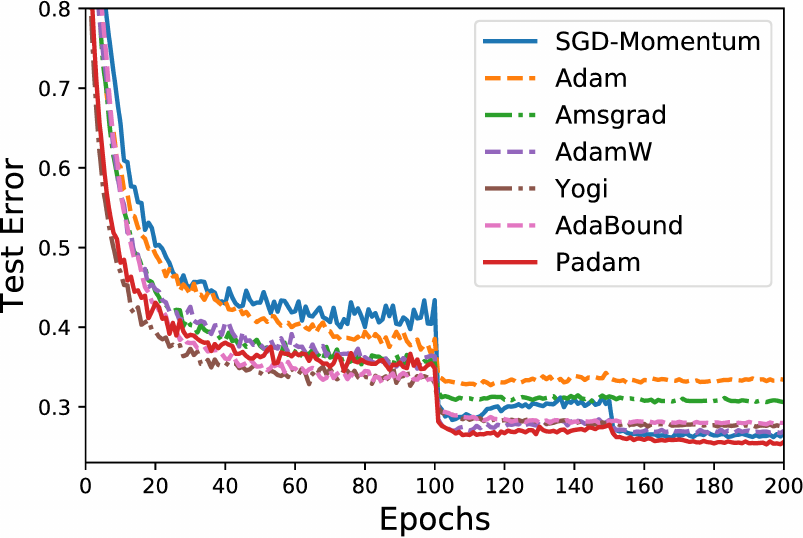}}
    \subfigure[Test Error for ResNet]{\includegraphics[width=0.32\textwidth]{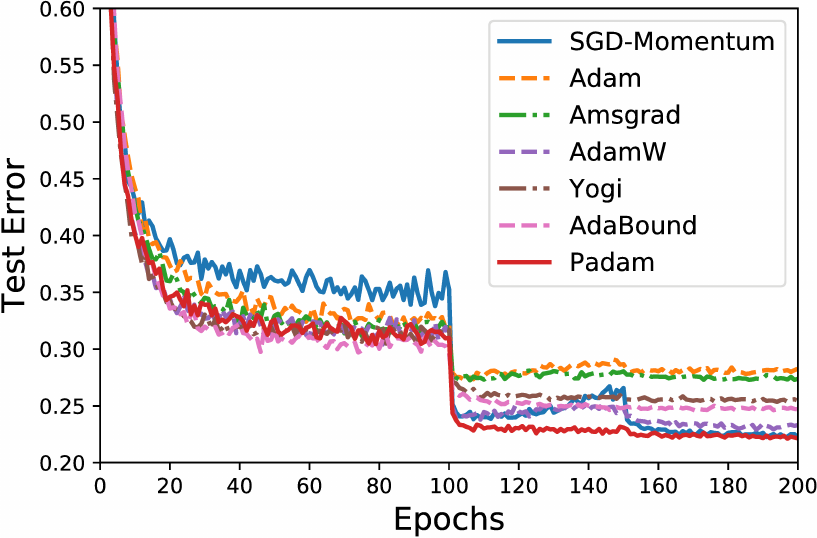}}
    \subfigure[Test Error for WideResNet]{\includegraphics[width=0.32\textwidth]{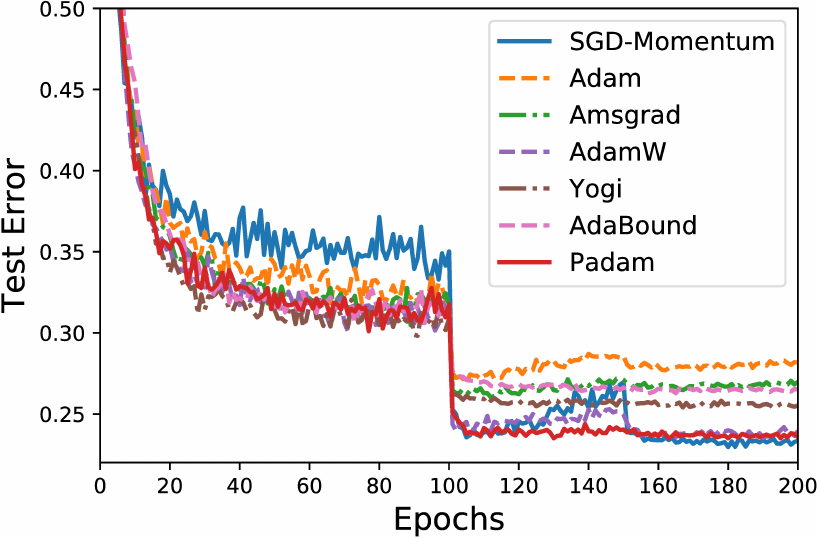}}
    \caption{Train loss and test error (top-$1$ error) of three CNN architectures on CIFAR-100. In all cases, Padam achieves the fastest training procedure among all methods and generalizes as well as SGD with momentum.}
    \label{fig:cifar100}
\end{figure*}
We conducted additional experiments using CIFAR-100 dataset.
Figure \ref{fig:cifar100} plots the train loss and test error (top-$1$ error) against training epochs on the CIFAR-100 dataset. The parameter settings are the same as the setting for CIFAR-10 dataset.
We can see that Padam achieves the best from both worlds by maintaining faster convergence rate while also generalizing as well as SGD with momentum in the end.

Table \ref{table:cifar100} show the test accuracy of all algorithms on CIFAR-100 dataset.
We can observe that Padam achieves the highest test accuracy in  VGGNet and ResNet experiments for CIFAR-100. On the other task (WideResNet for CIFAR-100), Padam is also on a par with SGD with momentum at the final epoch (differences less than $0.1\%$).

\begin{table}[ht!]
  \caption{Final test accuracy of all algorithms on CIFAR-100 dataset. Bold number indicates the best result.}
  \label{table:cifar100}
  \vskip 0.15in
  \begin{center}
%   \begin{small}
%   \begin{sc}
  \begin{tabular}{c|ccccccc}
    \toprule
    \multicolumn{1}{c}{\multirow{2}{*}{Models}} & \multicolumn{7}{c}{Test Accuracy (\%) }\\
    % \cline{2-5}
    \cmidrule(r){2-8}
    \multicolumn{1}{c}{} & SGDM & Adam & Amsgrad  & AdamW & Yogi & AdaBound & Padam\\
    \midrule
    VGGNet & $  73.32 $ & $  66.60 $ & $  69.40 $ & $  73.03 $ & $  72.35 $ & $  72.00 $ & $  \mathbf{74.39}$\\
    ResNet & $  77.77 $ & $  71.72 $ & $  72.62 $ & $  76.69 $ & $  74.55 $ & $  75.29 $ & $  \mathbf{77.85} $\\
    WideResNet & $  \mathbf{76.66} $ & $  71.83 $ & $  73.02 $ & $  76.04 $ & $  74.47 $ & $  73.49 $ & $  76.42 $ \\
    \bottomrule
  \end{tabular}
%    \end{sc}
%   \end{small}
  \end{center}
  \vskip -0.1in
\end{table}

\end{document}